\documentclass[journal]{IEEEtran}
\usepackage{amsmath,amsfonts}
\usepackage{newtxtext}
\usepackage{array}
\usepackage{textcomp}
\usepackage{stfloats}
\usepackage{verbatim}
\usepackage{enumitem}
\usepackage{graphics}
\usepackage{tikz}
\usepackage{tabularx}
\usepackage{listings}
\usepackage{enumitem}
\usepackage{makecell}
\usepackage{algpseudocode}% http://ctan.org/pkg/algorithmicx
\usepackage{amssymb}
\usepackage[all]{nowidow}

\usepackage[normalem]{ulem}
\usepackage{mathtools}
\usepackage{url}

\usepackage[utf8]{inputenc}
\usepackage{enumitem}
\usepackage{url}
\usepackage{tikz}
\usepackage{enumitem}
\usepackage{multirow}
\usepackage{longtable}
\usepackage{listings}
\usepackage{float}
\usepackage{comment}
\usepackage{hyperref}
\usepackage[ruled,lined,boxed,linesnumbered]{algorithm2e}
\usepackage{amsthm}

\usepackage{titlesec}
\titlespacing*{\section}
{0pt}{5pt}{3pt}
\titlespacing*{\subsection}
{0pt}{5pt}{3pt}
\titlespacing*{\subsubsection}
{0pt}{5pt}{3pt}
\usepackage{xcolor}
\usepackage{colortbl}
\colorlet{darkgreen}{green!80!black}
\definecolor{lightmagenta}{rgb}{1, 0.5, 1}
\definecolor{brown}{rgb}{0.6, 0.3, 0.0}
\definecolor{purple}{rgb}{0.5, 0.0, 0.5}
\definecolor{palered}{RGB}{255,106,106}

\newcommand{\sk}[1]{\textcolor{black}{#1}}

\newcommand{\blue}[1]{\textcolor{blue}{ #1}}

\newcounter{reviewercount}
\newcounter{commentcount}

\newcommand{\aeditor}%
  {\bigskip\noindent {\bf COMMENTS OF THE ASSOCIATE EDITOR}%
  \setcounter{commentcount}{0}\par 
}

\usepackage[utf8]{inputenc}
\usepackage{enumitem}
\usepackage{url}
\usepackage{graphicx}
\usepackage{tabularx}
\usepackage{listings}
\usepackage{float}
\usepackage{comment}
\usepackage[ruled,lined,boxed,linesnumbered]{algorithm2e}
\usepackage{soul}

\theoremstyle{plain}

\theoremstyle{definition}

\newtheoremstyle{claim}% name
  {\topsep}% space above
  {\topsep}% space below
  {}% body font
  {}% indent amount
  {\itshape}% theorem head font
  {}% punctuation after theorem head
  {.5em}% space after theorem head
  {\thmname{#1}\thmnumber{ #2}\thmnote{ (#3)}}% theorem head spec

\SetKwInput{KwOutput}{Output}
\SetKwInput{KwInput}{Input}

\usepackage{flushend}

\begin{document}
\setcounter{page}{1}
\twocolumn

\title{An Updated Assessment of Reinforcement Learning for Macro Placement}

\author{
    Chung-Kuan Cheng,~\IEEEmembership{Fellow,~IEEE}, 
    Andrew B. Kahng,~\IEEEmembership{Fellow,~IEEE}, 
    Sayak Kundu,~\IEEEmembership{Student Member,~IEEE},
    Yucheng Wang,~\IEEEmembership{Student Member,~IEEE},
    and Zhiang Wang,~\IEEEmembership{Student Member,~IEEE}
}

\maketitle

% \vspace{-1in}

%\begin{document}

\begin{abstract}
 
We provide an improved assessment of Google Brain's deep reinforcement 
learning approach to macro placement \cite{MirhoseiniGYJSWLJPNPTSHTLLHCD21} 
and its updated Circuit Training ({\em CT}) implementation in 
GitHub \cite{CT-TCAD}.
A stronger simulated annealing ({\em SA}) baseline leverages the
``go-with-the-winners'' metaheuristic~\cite{gwtw} and 
a multi-threading implementation.
We develop and release new public benchmarks in sub-10nm technology: LEF/DEF 
for Google's 7nm TSMC Ariane protobuf and scaled variants, as well as
testcases implemented in the open-source ASAP7 7nm research enablement.
We evaluate from-scratch training and fine-tuning results
for the latest ``AlphaChip'' release of Circuit Training, alongside multiple
alternative macro placers. 
We also study the recently-published pre-training guidance in \cite{CT-TCAD}. 
A commercial place-and-route tool is used to provide ``true reward''
post-route \sk{power, performance and area} metrics. All data,
evaluation flows and related scripts are publicly available in the
{\em MacroPlacement} GitHub repository \cite{TILOS-repo}. 
Our study affords insights into reproducibility and reporting
in the research literature, and points out still-missing 
confirmations (e.g., of {\em CT}'s scalability and pre-training methodology)
that remain open questions for the research community.
\end{abstract}

\section{Introduction}
\label{sec:intro}

Macro placement is a fundamental problem in VLSI physical design that 
involves determining the positions of large circuit blocks (macros) on 
a chip layout canvas. These macros typically include memory arrays, processor 
cores, analog blocks, and other pre-designed components that are significantly 
larger than standard cells. The macro placement problem is NP-hard and involves 
complex trade-offs between multiple objectives including wirelength, area 
utilization, timing closure, power consumption, and routing 
congestion. The quality of macro placement directly impacts the final chip 
performance, manufacturability, and cost, making it one of the most critical 
steps in modern chip design flows.

In June 2021, authors from the Google Brain and Chip Implementation and 
Infrastructure (CI2) teams reported a novel reinforcement learning (RL) 
approach for macro placement \cite{MirhoseiniGYJSWLJPNPTSHTLLHCD21} ({\em Nature}). 
The authors stated, ``In under six hours, our method automatically
generates chip floorplans that are superior or comparable to those produced by
humans in all key metrics, including power consumption, performance and chip area.''
Results were reported to be superior to those of the RePlAce academic 
placer \cite{ChengKKW18} and the simulated annealing ({\em SA}) metaheuristic.

{\em Nature} authors promised data and code availability, e.g.,
``The code used to generate these data is available from the corresponding 
authors upon reasonable  request''. The Circuit Training ({\em CT}) 
repository \cite{CT-TCAD}, which ``reproduces the methodology published 
in the Nature 2021 paper'', was made public in January 2022.
However, reproduction and evaluation of {\em Nature} and {\em CT} 
has been hampered because neither the data nor the code used in these 
works is, \sk{as of November 2025}, {\em fully} available.
The irreproducibility of
{\em Nature} claims has led to controversy and slowed progress in the field.

This situation reflects a broader scientific concern with reproducibility
and integrity. When high-profile claims shape research agendas and public
perception, transparent and replicable evaluation becomes essential. % a first-order need.
Recent discussions in {\em Nature} (Jan.~2025)~\cite{Udesky25}
and work in {\em PNAS} (Aug.~2025)~\cite{RichardsonHBSA25} emphasize this point. For macro
placement specifically, broad coverage in mainstream and trade press of
RL-based approaches after {\em Nature} underscores the EDA community's
responsibility to deliver clear,
technically rigorous, and reproducible assessments of empirical merit and practical impact.

A 2022 ``Stronger Baselines'' ({\em SB}) study, performed
at Google \cite{StrongerBaseline}, claimed that a properly-implemented
simulated annealing outperforms {\em Nature}. 
\cite{StrongerBaseline} used a Google-internal version 
of {\em CT} with different benchmarks and evaluation metrics.
Our % \sout{subsequent invited}
conference paper \cite{ispd23} presented efforts toward 
an open-source %\brown{\sout{, transparent}} 
implementation and assessment of {\em Nature} 
and {\em CT}. \sk{We made the data, scripts and results} public in the 
{\em MacroPlacement} GitHub repository \cite{TILOS-repo}, with additional 
background, details, FAQs, etc. given in updates \cite{FtR}, an extended 
arXiv version \cite{ispd23arXiv}, and ``Our Progress'' \cite{OurProgress} 
and other documentation in \cite{TILOS-repo}.

In 2024, {\em Nature} authors published
an addendum~\cite{GoldieMYJSWLJPNPTSHTLLHCD24} to clarify 
methods and results of \cite{MirhoseiniGYJSWLJPNPTSHTLLHCD21}.
Importantly, updates to {\em CT}, dubbed ``AlphaChip'', were publicly 
released with pre-trained model weights (CT-AC) \cite{CT-TCAD}.\footnote{Below,
we refer to training Circuit Training (commit hash: 4c6fd98) 
from scratch as {\em CT-Scratch}. We denote fine-tuning of ``AlphaChip'' using
the checkpoint released in August 2024 (commit hash: 4c6fd98) as {\em CT-AC}.}
Markov \cite{Markov24} published a peer-reviewed meta-analysis of the
{\em Nature} results, cross-checking \cite{StrongerBaseline} and \cite{ispd23}.
Goldie et al., in the arXiv post \cite{GoldieMD24}, criticized \cite{ispd23},
raising issues such as: (i) absence of peer-review; 
% (ii) no pre-training of  RL in the reported studies; 
(ii) lack of RL pre-training;
(iii) potential non-convergence of training for some
testcases; (iv) \sk{use of 45nm and 12nm technologies versus sub-10nm 
(TSMC 7nm, corresponding to TPU v4) in {\em CT}}; and (v) insufficiency 
of compute resources used.

Against this backdrop, rather than proposing a new algorithm, 
we carefully evaluate the extent to which the {\em Nature}
method improves upon {\em prior approaches}, using extensive experimentation
and a transparent methodology. In this work, we re-execute and augment the
studies reported in \cite{ispd23} to thoroughly and conclusively address
the criticism in \cite{GoldieMD24}. Through this effort, we obtain a more
rigorous assessment of the {\em CT} approach, along with further insights
into reporting and reproducibility in the research literature. Our main
contributions beyond \cite{ispd23} are as follows.

\begin{itemize}[noitemsep,topsep=0pt,leftmargin=*]
\item {\bf Evaluation of updated CT.} For all testcases, we 
% \brown{\sout{now}} 
train the updated Circuit Training from 
scratch, and also fine-tune AlphaChip from Google's August 2024
pre-trained checkpoint. Experimental observations, including resource
requirements, model quality, and convergence and variability behavior,
are reported in Sections \ref{sec:experiment} and \ref{sec:pretrain} below. 

\item {\bf Improved SA.} We strengthen the simulated annealing baseline
by incorporating 
multi-threading and a 1994 ``go-with-the-winners'' \cite{gwtw} metaheuristic, 
while also ensuring reproducibility of executions (see Section
\ref{sec:sa}). The improved {\em SA} achieves up to 26\% better
proxy cost within the same runtime while using 
only a quarter of the resources, compared to the {\em SA}
implementation reported in \cite{ispd23}. \sk{The improved SA
maintains superior performance over recent {\em CT-AC} methods,
reaffirming the effectiveness of carefully optimized classical
heuristics for combinatorial optimizations in the context of
macro placement.}

\item {\bf Sub-10nm testcases.} 
We strengthen sub-10nm experimental
enablement in two ways. (i) We convert Google's public TSMC 7nm
Ariane testcase ({\em CT-Ariane}) from protobuf \cite{CT-TCAD} to
LEF/DEF; we publish this and 
additional scaling studies of macro placement optimizers,
following the ``quantified scaling suboptimality'' 
methodology of \cite{HagenHK95} and evaluating AlphaChip's 
performance on ``blocks with over 500 macros'' 
\cite{GoldieMYJSWLJPNPTSHTLLHCD24} in the {\em MacroPlacement}
repository \cite{TILOS-repo}.
(ii) We port our testcases to a second 7nm enablement, the open-source 
academic ASAP7 PDK \sk{from}
% originally developed by 
ASU/Arm~\cite{ASAP7}. Studies 
with these new sub-10nm enablements \sk{reaffirm findings} of \cite{ispd23}.

\item {\bf Pre-training studies.} We perform % \brown{\sout{our own}}
pre-training of {\em CT} following the instructions 
in the Circuit Training repository \cite{CT-TCAD}.
Our studies (see Section \ref{sec:pretrain}) highlight the need for further
confirmations of scalability, resource efficiency and other claims
in \cite{MirhoseiniGYJSWLJPNPTSHTLLHCD21}.

\item {\bf Addressing resources and convergence.} 
Our revised experimental protocol provides
compute resources that are sufficient for {\em CT} per 
\cite{MirhoseiniGYJSWLJPNPTSHTLLHCD21} \cite{CT-TCAD}.
Furthermore, in both training from scratch ({\em CT-Scratch})
and fine-tuning ({\em CT-AC}) -- for all our testcases
-- we \sk{double} iterations from 200 to 400 to 
provide sufficient opportunity for {\em CT} to converge, and
conduct multiple trials before declaring non-convergence.

\end{itemize}

Our updated and strengthened evaluation reconfirms
conclusions of \cite{ispd23}. Our contribution is evidence-based evaluation --
rigorous experiments, strengthened baselines, and reproducible sub-10nm
enablements -- rather than a new placement algorithm; this prioritizes clarity
on empirical merit and practical impact. 
The simulated annealing and human baselines continue to show
superiority to the latest AlphaChip 
while using substantially fewer resources. Further, {\em scaled} sub-10nm 
Ariane variants expose \sk{additional weaknesses of
{\em Nature}}, e.g., regarding stability, scalability, and resource demands.
The {\em MacroPlacement} 
effort highlights the importance of ``frictionless reproducibility''~\cite{Donoho24}, 
along with open source code and data releases ``upon which others then build'' 
\cite{SpectorN18}, in the academic EDA field and its nexus with AI/ML.

In the following, Section \ref{sec:methods} lists the macro placement
methods studied, Section \ref{sec:replication} describes efforts toward open-
source replication of {\em CT} and Section~\ref{sec:sa} details our simulated
annealing approach. Section \ref{sec:benchmarks_and_flow} presents our experimental
setup, and Sections \ref{sec:experiment} and \ref{sec:pretrain} present results.
Section \ref{sec:conclusion} provides conclusions and directions for 
future research.

\section{Macro Placement Methods}
\label{sec:methods}

VLSI physical design researchers and practitioners have studied macro placement 
for well over half a century, as reviewed in \cite{MarkovHK15} \cite{WangCC09}. 
In this work, we study the following macro placement methods.
\begin{itemize}[noitemsep,topsep=0pt,leftmargin=*]
\item \textbf{Circuit Training} \cite{CT-TCAD} uses the RL approach to 
sequentially place macros. 
{\em CT} first divides the layout canvas into small grid cells, then uses
placement locations along with hypergraph partitioning
to group standard cells into standard-cell clusters (soft macros), to 
set up the environment.
The RL agent then places macros one by one onto the centers of grid cells; 
after all macros are placed, force-directed placement is used to 
determine the locations of standard-cell clusters and calculate the {\em proxy cost}.
Proxy cost is comprised of three metrics -- wirelength, density, and congestion -- 
which are proxies for routed wirelength, design density, and
routing congestion. Lower values of these metrics imply better design
quality.\footnote{\label{fn:proxycost}\cite{ispd23} provides a detailed description of
these proxy cost components. An open-sourced
implementation~\cite{ProxyCost} reproduces the black-box
implementation of the proxy cost in {\em CT}.}
Finally, the negative of the {\em proxy cost} is provided as the reward feedback
to the RL agent.

In this work, we use three variants of Google's RL approach:
(i) training AlphaChip (i.e., the latest version of the renamed framework
in the Circuit Training repository \cite{CT-TCAD}) from scratch 
(denoted as {\em CT-Scratch}); (ii) fine-tuning AlphaChip using the 
pre-trained checkpoint released in August 2024 (denoted as {\em CT-AC});
and (iii) fine-tuning AlphaChip using our own checkpoint pre-trained with specific
testcase variants (denoted as {\em CT-Ours}). Note that
our present work uses  Circuit Training~\cite{CT-TCAD} 
commit hash 4c6fd98 from February 2025, while our 
previous work~\cite{ispd23} used commit hash 91e14fd from August 2022.

% In this work, we use two variants of the RL approach:
% training Circuit Training from scratch (noted as CT)
% and fine-tuning AlphaChip using the checkpoint relased 
% in August 2024 (noted as CT-AC).

% In this work, we use three variants of the RL approach:
% training Circuit Training from scratch (noted as CT-Scratch), 
% fine-tuning AlphaChip using the checkpoint released 
% in August 2024 (noted as CT-AC), and fine-tuning AlphaChip 
% using our checkpoint pre-trained with netlist variants 
% (noted as CT-Ours).

\item \textbf{RePlAce} \cite{ChengKKW18} \cite{RePlAce} models the layout and netlist as an 
electrostatic system. \sk{Instances are modeled as electric charges, and the
density penalty as potential energy.} Instances are spread 
apart according to the gradient with respect to the density penalty.
Note that our present work uses RePlAce from OpenROAD \cite{OpenROAD-Ajayi-DAC19} \cite{RePlAce}, 
commit hash f02a3d4 from August 2024, which is the appropriate comparison; 
our previous work~\cite{ispd23} used a specific standalone
RePlAce chosen to match the ``Stronger Baselines'' study \cite{RePlAce2},
and {\em Nature} used a standalone RePlAce from the OpenROAD project
repository \cite{PeerReviewNature}~\cite{CT-RePlAce}, which was
deprecated in January 2021.
%RePlAce's underlying formulation and algorithm in OpenROAD are the same as in the variants studied by \cite{StrongerBaseline} \cite{MirhoseiniGYJSWLJPNPTSHTLLHCD21}.

% \item \textbf{AutoDMP}~\cite{AutoDMP} from Nvidia builds on the 
% GPU-accelerated global placer DREAMPlace \cite{DREAMPlace} and 
% detailed placer ABCDPlace \cite{ABCDPlace}.
% AutoDMP adds enhanced concurrent macro and standard cell placement,  
% along with automatic parameter tuning based on multi-objective 
% Bayesian optimization (MOBO).\footnote{In this work, 
% we do not include the results of {\em AutoDMP}.
% The experimental results of {\em AutoDMP} are in \cite{ispd23}.}

\item \textbf{CMP} is a state-of-the-art commercial macro placer from Cadence 
Design Systems, which performs concurrent macro and standard-cell placement. 
CMP has been available in Innovus place-and-route tool from 2019 versions
onward. CMP results also serve as input to the Cadence Genus iSpatial 
physical synthesis tool. We include results of CMP in our experimental study
(see Section \ref{sec:experiment}).

\item \textbf{Human-Expert} macro placements are contributed by individuals
at IBM Research \cite{jjung22}, ETH Zurich and UCSD 
\cite{Matheus}; human-expert placements are one of the two
baselines used by {\em Nature} authors \cite{MirhoseiniGYJSWLJPNPTSHTLLHCD21}.

\item \textbf{Simulated Annealing ({\em SA})} is the second baseline used 
by {\em Nature} authors, and is studied by both {\em Nature} and {\em SB}.
Annealing is applied to place macros in the same grid cells as 
{\em CT} (see Section \ref{sec:sa}).  
\end{itemize}

All testcases and results from the above methods, along with all scripts
and codes where applicable, are publicly available
in the {\em MacroPlacement} GitHub repository 
\cite{TILOS-repo}.\footnote{\label{fn:two}{\em MacroPlacement}~\cite{OurProgress}
also includes macro placement solutions from AutoDMP~\cite{AutoDMP} and 
Hier-RTLMP~\cite{hierRTLMP}. In this work, we do not compare 
with AutoDMP and Hier-RTLMP, as these were released after the
{\em Nature} work.} Based on permission from Cadence Design
Systems, we are able to make public the Cadence runscripts
used to obtain post-route \sk{power, performance and area (PPA)} metrics
 \sk{-- i.e., final chip metrics --} from macro placement solutions. Licensed users of Cadence Genus 21.1 
and Cadence Innovus 21.1 are able to fully replicate our results, through
post-route PPA metrics.\footnote{Note that the CT ``grouping''
step is implemented using the hMETIS binary, which is nondeterministic.
Therefore, clustered netlists used in \cite{ispd23} are not
reproducible, nor are studies of CT/AlphaChip that run the clustering
step on a gate-level netlist. We describe below how use of the hMETIS 
shared library can avoid this nondeterminism (see Subsection~\ref{subsec:ablation}).}

Last, we note that since 2021, many researchers from the machine learning 
and EDA communities have proposed various RL-based macro placement 
methods \cite{ChengY21, ChengLLYHY22, GuGPZXGY24, GengWLXT24, GuGLZ23, LaiML22, LaiLTWHL23, TanM24, YaoLL25, WangZTZ24, ZhaoYSTX23}. However, these 
works only show results on (non-real, old-node) physical design contest 
testcases (a practice that has drawn criticism from Google 
authors \cite{GoldieMD24}), and do not report post-route PPA metrics. 
In \cite{LeNBKLKKJK23}, the authors combine Circuit Training with 
simulated annealing to handle rectilinear layouts, and further show 
the results on proprietary in-house testcases. 
Our ongoing outreach to the authors in this recent literature aims to 
draw more attention to {\em MacroPlacement} testcases, runscripts 
and evaluation methods, to spur further assessment and understanding
of the RL approach.

%%% 
% @Sayak, do we need to mention here the computing resources needed for reproducing our results ?
%\textcolor{blue}{TODO:   discuss reproducibility of results (include what is needed for that) -- Cadence, our results are repeatable...}

\section{Replication of Circuit Training}
\label{sec:replication}

In this section, we discuss clarifications and reproduction
in open source of {\em CT}.
We first summarize key mismatches between {\em CT} and {\em Nature}. 
We then discuss the computing resources needed for 
studies of Circuit Training. 
As in \cite{ispd23}, we are thankful to Google engineers for 
answering questions and for many discussions that helped our 
understanding of {\em CT} starting in April 2022.
%Finally, we present updates in Circuit Training from 2022.

\subsection{Discrepancies between {\em CT} and {\em Nature}}
\label{subsec:mismatch}

From its outset, the {\em CT} GitHub repo has been stated to reproduce 
the methodology published in {\em Nature} \cite{CT-Claim}.
Yet, several discrepancies
between {\em CT} and the claims of {\em Nature} 
authors should be noted in the present context 
\cite{ispd23} \cite{ispd23arXiv}.

\begin{itemize}[noitemsep,topsep=0pt,leftmargin=*]

\item {\em Availability of code.} Two key ``blackbox'' 
  elements, i.e.,  force-directed placement and proxy cost calculation, 
  are neither clearly documented in {\em Nature} nor visible in {\em CT}. 
  Reverse-engineering and replication in open-sourced C++ are discussed 
  in \cite{ispd23} and Section \ref{sec:sa} below. 

  \item {\em Need for pre-training.} {\em Nature} 
  \cite{MirhoseiniGYJSWLJPNPTSHTLLHCD21} does not show 
  benefits from pre-training in its ``Table 1'' metrics. Rather,
  \cite{MirhoseiniGYJSWLJPNPTSHTLLHCD21} only shows benefits (from 
  the pre-trained model) in terms of runtime and final proxy cost. 
  Additionally, the January 2022 ARIANE.md in Circuit 
  Training \cite{Ariane-pre-train} states that ``Our results 
  training from scratch are comparable or better than the reported results in the paper (on page 22) which used fine-tuning from a pre-trained model''.
  
   \item {\em Gridding of macro placement locations.}  The method described
   in {\em Nature} ``place[s] the centre of macros and standard cell 
   clusters onto the centre of the grid cells''. However, {\em CT} does 
   not require standard-cell clusters to be placed onto centers of 
   grid cells. 
  
   \item {\em Adjacency matrix construction.}
   The {\em Nature} paper describes generation of the adjacency 
   matrix based on the {\em register distance} between pairs of nodes. 
   This is a well-known technique (e.g., \cite{VidalCP20}) that is 
   consistent with timing being a key metric for placement quality. 
   However, {\em CT} builds its adjacency matrix based only on direct 
   connections between nodes (i.e., macros, IO ports and standard-cell clusters).
   % XXX Is the above summary correct?
\end{itemize}

For completeness, we recognize that approximately 2.5 years of effort and
updates are embodied in the delta between {\em CT}'s commit hash 91e14fd studied 
in \cite{ispd23} and commit hash 4c6fd98 that we study here \cite{CT-TCAD}.
This delta has brought numerous changes to {\em CT}, spanning
functionalities, default settings, and library dependencies.  For example,
(i) {\em CT} has now open-sourced pre-training and fine-tuning scripts, which
we use in our present study. (ii) {\em CT} has also enabled use of DREAMPlace 
\cite{DREAMPlace} to finalize soft macro placement. 
However, since our goal is to evaluate the claims in the original {\em Nature}
paper -- not newer variations that mix RL with other techniques already known
to work well for macro placement -- we do not use the DREAMPlace-enabled {\em CT}
in our experiments.
(iii) Parameter changes between the two commit hashes include 
reduction of gradient clipping from 1.0 to 0.1; reduction of the
penalty for infeasible placement from -1 to -4; and reduction of 
\#episodes per iteration from  1024 to 256 (which is stated to help
with convergence and the final return). (iv) The hidden binary
for {\em plc\_client} has been updated from version 0.0.3 to 0.0.4,
and our environment hence updates several library dependencies: 
TF-Agent 0.14.0 to 0.19.0, TensorFlow 2.10.0 to 2.15.0, 
dm-reverb 0.9.0 to 0.14.0, and CUDA 11.8 to 12.2.
    
\subsection{Computing resources for Circuit Training}
\label{subsec:compute_resource}

A critical concern \cite{GoldieMD24} is whether compute resources 
are sufficient to enable assessment of Circuit Training and reproduction of 
the {\em Nature} results.
Such resources have three main dimensions: training server;
collect servers; and training iterations (equivalently, steps or walltime).
To assess adequacy of resources, we rely on documentation from
Google authors in, e.g., \cite{MirhoseiniGYJSWLJPNPTSHTLLHCD21} \cite{YuSJBGMG22} \cite{CT-TCAD}.

\noindent
{\bf Training server.} 
We use eight NVIDIA-V100 GPUs to train the model for global batch size = 1024.
We believe this is adequate, based on \cite{YuSJBGMG22} where the authors state,
``We think the 8-GPU setup is able to produce better results primarily
because it uses a global batch size of 1024, which makes learning more
stable and reduces the noise of the policy gradient estimator. Therefore,
we recommend using the full batch size suggested in our open-source
framework in order to achieve optimal results.''
Circuit Training \cite{CT-TCAD} itself shows the use of an 8-GPU setup to 
reproduce their published Ariane results \cite{Ariane-pre-train}.
    % \item 
    % \textcolor{blue}{@SY and Yucheng,  please update this paragraph if necessary.}
The global batch size = 1024 used in our runs is the same global batch 
size that is used in the Nature paper 
\cite{MirhoseiniGYJSWLJPNPTSHTLLHCD21}.\footnote{\cite{MirhoseiniGYJSWLJPNPTSHTLLHCD21} refers
to the use of 16 GPUs. However, based on the statements 
in \cite{YuSJBGMG22} and what Circuit Training describes for ``reproduce 
results'', the final proxy cost achieved by our environment should not 
differ materially from the environment with 16 GPUs described 
in \cite{MirhoseiniGYJSWLJPNPTSHTLLHCD21}. Indeed, using our setup we 
achieve similar proxy cost for Google's Ariane testcase (CT-Ariane) as 
reported in {\em CT} (see Subsection~\ref{subsec:comparison}).}
% \end{itemize}

\noindent
\textbf{Collect servers.} 
In \cite{YuSJBGMG22}, Google authors write that ``with distributed 
collection, the user can run many  (10s-1000s) Actor workers with each 
collecting experience for a given policy, speeding up the data 
collection process.” They further explain that ``as mentioned in 
Section 2.2, data collection and multi-GPU training in our framework 
are independent processes which can be optimized separately.''

Our previous work \cite{ispd23} used two collect servers each running 
13 collect jobs, i.e., a total of 26 collect jobs were used for data 
collection. In our present work, we increase this to five collect 
servers each running  at  least 51 collect jobs, i.e., a total of 256 
collect jobs are used for  data collection. 
We believe this is adequate, again based on
\cite{YuSJBGMG22}, which suggests that increasing collect jobs 
beyond this point has diminishing returns. (The {\em Nature} authors 
run 512 collect jobs for data collection, with the number of collect servers 
used to run these 512 collect jobs being unclear from the description provided. 
At the same time, \cite{YuSJBGMG22} indicates that a larger number of collect 
jobs only speeds up training without affecting the outcome quality.)
Since we use fewer collect jobs, our runs are slower, but quality is not 
compromised. We expect our runtimes to be higher than what Nature reports,
and we define experimental protocols accordingly, as described next.

\noindent 
{\bf Training iterations.} Published Google materials indicate % allow us to infer
a sufficient number of training iterations to use for {\em CT} 
in our experiments.

\begin{figure}[ht]
    \centering
    \includegraphics[width=0.45\columnwidth]{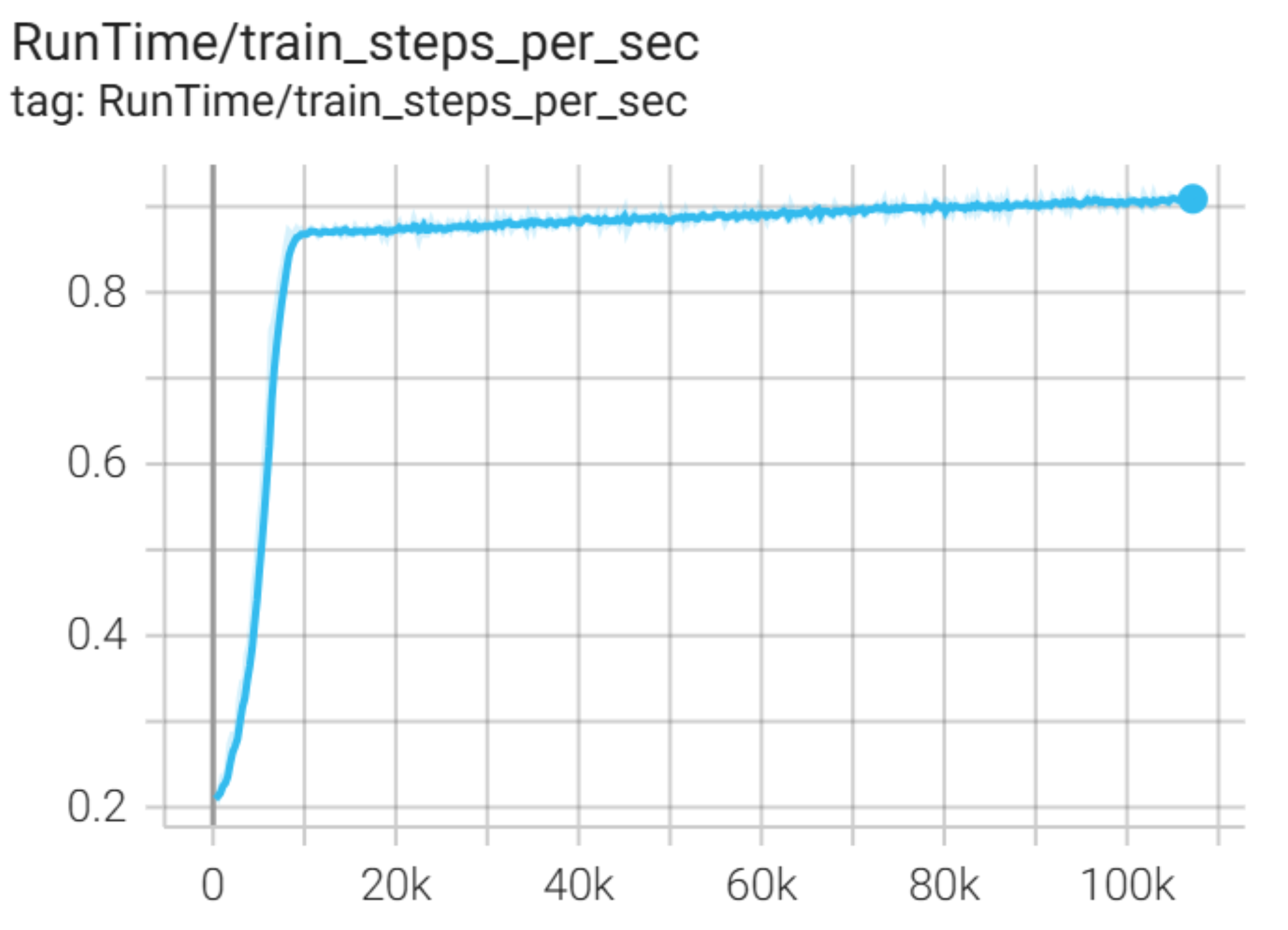}
    \includegraphics[width=0.45\columnwidth]{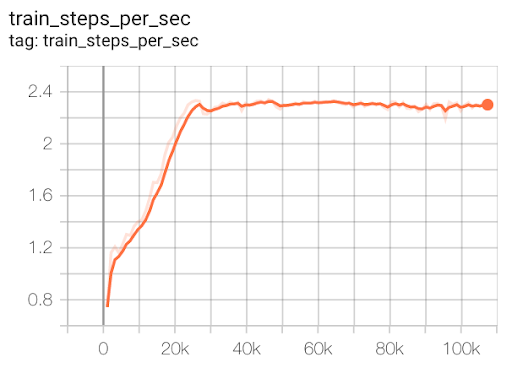}
    \vspace{-0.05in}
    \caption{Train steps per second plot for our CT run
    \sk{(left)} and the CT run available \sk{(right)} 
    in the CT repository for Ariane.}
    \label{fig:train_step}
\end{figure}

\begin{itemize}[noitemsep,topsep=0pt,leftmargin=*]
  \item 
  {\em Train steps per second} is the indicator of the {\em CT} training speed. 
    Figure \ref{fig:train_step}'s left plot shows the 
    {\em CT} training speed of $\sim$0.9 steps/sec
    for Ariane \cite{Ariane-pre-train} in our environment.
    The right plot shows the {\em CT} training speed 
    ($\sim$2.3 steps/sec)  for {\em CT-Ariane}  from a 
    TensorBoard (no longer available) in the 
    {\em CT} \cite{Ariane-pre-train} repo.
    From this, we infer that our runtime is expected to be approximately 
    2.6$\times$ longer in our environment, compared to when the 
    resources suggested in the {\em CT} repo are used.

    \item We observe that~\cite{Ariane-pre-train} gives 200 as a 
    suggested number of iterations.
    % , and we make this the default iteration budget.
    To ensure adequacy of iterations afforded to {\em CT}
    in our experiments, 
    % whenever we 
    % see a failure to converge after the default 200 iterations, 
    we provide {\em another} 200 iterations, for a total of 400.  
    Moreover, for larger testcases (BlackParrot Quad-Core 
    and MemPoolGroup; see Table \ref{tab:testcase}),
    % we increase the default number of training iterations 
    % of both {\em CT-Scratch} and {\em CT-AC} to 400.
    we extend further: in light of non-determinism in {\em CT} behavior, 
    if training fails to converge after 400 iterations, we execute up 
    to two additional 400-iteration runs before declaring 
    non-convergence (i.e., ``Divergence'' in Tables \ref{tab:mp_result} and \ref{tab:ct_ariane}). 
\end{itemize}

\section{A Stronger Annealing Baseline}
\label{sec:sa}

Both {\em Nature} and {\em SB} use simulated annealing
({\em SA}) \cite{KirkpatrickGV83} as a baseline for comparison. 
Table 2 of \cite{StrongerBaseline} gives a concise comparison of 
hyperparameters used by the two works. 
As in \cite{ispd23}, we implement and run {\em SA} based on the description 
given in the {\em SB} manuscript. Our implementation differs from that 
described in {\em Nature} in its use of {\em move} and {\em shuffle} 
in addition to {\em swap}, {\em shift} and {\em mirror} actions. 
We also use two initial macro placement schemes, 
i.e., ``spiral macro placement'' whereby macros are sequentially 
placed around the boundary of the chip canvas in a counterclockwise spiral 
manner, and ``greedy packer''  whereby macros are packed in sequence 
from the lower-left corner to the top-right corner of the chip canvas 
\cite{StrongerBaseline}. 
Force-directed (FD) placement is used to update the locations of 
standard-cell clusters every $\{2n, 3n, 4n, 5n\}$ macro actions, where 
$n$ is the number of hard macros; FD is not itself an action. 
The {\em SA} cost function is the proxy cost, which consists of wirelength, 
density and congestion. 

As described in \cite{ispd23}, Google's implementations of
FD and proxy cost calculation are not open-sourced, but are
only available via the {\em plc\_client} in \cite{CT-TCAD}.
For speed and transparency, our {\em SA} experiments use our own
C++ reimplementations of FD and proxy cost calculation; however,
the Google {\em plc\_client} is used for final FD soft macro 
placement and proxy cost evaluation at the end of the {\em SA} run. 
(Section 3.2 in \cite{ispd23} provides details of 
force-directed placement and proxy cost congestion.) 
Our {\em SA} codes are open-sourced in {\em MacroPlacement} \cite{SA}.

% In our {\em ISPD-23} work, 
\noindent 
{\bf A Go-With-the-Winners Enhancement.} Similar to 
{\em Nature} and {\em SB}, our previous work in \cite{ispd23} runs 
320 {\em SA} workers in parallel for 12.5 hours. Each worker, with its own 
hyperparameter setting, operates independently and does not communicate 
with other workers. Then, the macro placement solution with minimum 
proxy cost (as calculated by our C++ code) is used as the final {\em SA} solution.
% and its post-FD proxy cost as evaluated by the {\em plc\_client} is returned as the SA proxy cost. 
However, running 320 {\em SA} workers in parallel requires multiple servers,
which may be impractical for users with limited computing resources.

To obtain a stronger {\em SA} baseline, we adopt the ``go-with-the-winners'' 
(GWTW) scheme~\cite{gwtw} in a multi-threading implementation. In essence, 
GWTW allows a set of solution threads to proceed independently, but 
periodically executes a `sync-up' whereby (i) the best threads are identified,
(ii) their solutions  are cloned to fill up the entire set of threads, and
(iii) the threads then independently continue the solution process until 
the next `sync-up'. This approach has seen previous adoption in physical 
design, e.g., for gate sizing \cite{HuKKKM12}.

The detailed algorithm is shown in Algorithm~\ref{algo:sa}.
The algorithm can be divided into following steps:
\begin{itemize}[noitemsep, topsep=0pt, leftmargin=*]
    \item {\bf Lines 1-10:} We initialize {\em SA} workers in parallel, with each using
    a unique random seed to shuffle same-size macros. 
    Placement is initialized via spiral initialization (resp.
    greedy packing) for workers with odd (resp. even) IDs.

    \item {\bf Lines 14-17:} Each {\em SA} worker is run for $sync\_iter$ iterations in parallel.
    $sync\_iter$ is set based on $sync\_freq$. We use $sync\_freq = 0.1$, meaning
    there are 9 synchronizations among the workers.
    
    \item {\bf Lines 18-21:} The algorithm stops when $Iter$ iterations are performed;
    otherwise, $syncWorkers$ selects the top $k$ workers based on proxy 
    cost and replicates their macro locations and orientations to the
    remaining workers evenly.
    %based on the replication ratio $r$.
    
    \item {\bf Line 22:} writes out the best macro placement solution in terms of
    proxy cost for each worker.
\end{itemize}

\vspace{-0.1in}
\begin{algorithm}[]
% \vspace{-0.05in}
\caption{Simulated Annealing}
\small
\label{algo:sa}
\KwIn{%
    \begin{tabular}[t]{@{}l@{}}
    % Move ratios: (swap, shift, flip, move, shuffle) $MR$, \\ 
    % {0.24, 0.24, 0.04, 0.24, 0.24
    Random seeds: $seed = 1$, \\
    Number of iterations: $Iters$, \\
    $N \times \#macro$ moves per iteration ($N = 20$), \\
    Initial temperature: $T_0 = 0.005$, \\
    Minimum temperature: $T_{min} = 1\times10^{-8}$, \\
    Cooling rate: $\alpha = \exp\Big(\frac{\ln(T_{min}/T_0)}{Iters}\Big)$, \\
    % Density weight: $\gamma = 0.5$, \\
    % Congestion weight: $\lambda = 0.5$, \\
    Number of workers: $W = 80$, \\
    Replicated top $k = 8$ workers, \\
    %Replication ratio: $r = [10, 10, \dots, 10]$, \\
    Synchronization frequency: $sync\_freq = 0.1$ \\
    % Enable fast synchronization: $is\_async = 0$
    \end{tabular}
}
\KwOut{Macro placement solutions.}
\BlankLine
\SetAlgoNoEnd

% \tcp{--- 1. Create and initialize workers in parallel ---}
$workers \leftarrow$ \text{create} $W$ \text{workers}\;
\For{$i \gets 0$ \textbf{to} $W-1$ \textbf{in parallel}}{
    %\tcp{Assign hyperparameters and seed for the $i$th worker}
    $workers[i].seed \gets seed + i$\;  $workers[i].N \gets N$\;
    $workers[i].T \gets T_0$\; $workers[i].\alpha \gets \alpha$\;
    %\tcp{Initialize macro placements differently based on parity of $i$}
    \If{$(i \bmod 2) = 0$}{
        $workers[i].\text{macro\_placement} \gets \text{``spiral macro placement''}$\;
    }
    \Else{
        $workers[i].\text{macro\_placement} \gets \text{``greedy packer''}$\;
    }
}

\BlankLine
%\tcp{--- 2. Define synchronization interval and iteration counters ---}
$\text{iter\_count} \gets 0$\;
$\text{sync\_iter} \gets Iters \times \text{sync\_freq}$\;

\BlankLine
\While{\text{true}}{
    %\tcp{Compute the iteration index up to the next synchronization point}
    $\text{end\_iter} \gets \min(Iters,\ \text{iter\_count} + \text{sync\_iter})$\;
    
    %\tcp{--- 3. Run each worker for the designated block of iterations in parallel ---}
    \For{$i \gets 0$ \textbf{to} $W-1$ \textbf{in parallel}}{
        %\text{run\_SA}\bigl(workers[i],\ \text{iter\_count},\ \text{end\_iter}\bigr)\;
        Each worker performs $(\text{end\_iter} - \text{iter\_count})$ {\em SA} iterations; applying $N \times \#macro$ moves per iteration and updating temperature\;
    }
    
    %\tcp{Update global iteration counter}
    $\text{iter\_count} \gets \text{end\_iter}$\;
    
    %\tcp{Check if we have reached the total number of iterations}
    \If{$\text{iter\_count} = Iters$}{
        \textbf{break}\;
    }
    
    \BlankLine
    %\tcp{--- 4. Synchronize: replicate top-$k$ solutions among the workers ---}
    $\text{candidate\_solutions} \gets \text{extractTopK}(workers,\ k)$\;
    Evenly distribute these top-$k$ solutions across all the workers\;
    %(according to replication ratio $r$)\;
    %\text{syncWorkers}\bigl(workers,\ \text{candidate\_solutions},\ r\bigr)\;
}

\BlankLine
\textbf{Write out the best solution of each worker.}
\end{algorithm}

As noted above, after placing soft macros (standard-cell clusters) with
GWTW {\em SA}, we use {\em CT}'s {\em plc\_client} to evaluate
the proxy cost of the best macro placement solutions for each 
worker, and then return the best solution in terms of proxy cost 
for P\&R evaluation. In our runs, the probabilities for five 
solution move operators (swap, shift, move, shuffle and flip) 
are respectively set to 0.24, 0.24, 0.24, 0.24 and 0.04. The number of 
iterations is set to ensure that overall runtime for each testcase is 
less than 12 hours on our slowest CPU server.\footnote{For Ariane, 
BlackParrot, MemPoolGroup, Ariane-X2 and Ariane-X4, we set
the number of iterations to 18K, 9K, 4.5K, 9K and 4K, respectively. 
This corresponds, e.g., to $\sim$11 hours on an Intel Xeon Gold 6148 CPU, 
or $\sim$3 hours on an AMD EPYC 9684X CPU.} 

Relative to the {\em SA} implementation in \cite{ispd23}, our present
{\em SA} implementation achieves similar or better results while using only 
one-fourth of the CPU resources: 80 threads instead of 320 threads, enabling
execution on a single CPU server. Further, to ensure exact reproducibility 
across different platforms, (i) we use lookup tables for exponent computation 
and provide a binarized version of the lookup table in our repository; and
(ii) we provide scripts to generate Docker and Singularity images that
reproduce the same environment. Our testing across a range of Intel Xeon Gold 
and AMD EPYC CPUs confirms exact matching of {\em SA} solutions obtained by
all 80 workers using the same Docker or Singularity image.

\section{Experimental Setup}
%Modern Benchmarks and Commercial Evaluation Flow}
\label{sec:benchmarks_and_flow}

We now describe our experimental setup.
We first describe testcases and design enablements.
%that have been developed to improve academic research foundations while also serving the {\em MacroPlacement} effort. 
We then present the commercial evaluation flow used 
to evaluate macro placement solutions.
Last, we present our settings for {\em CT}.

\subsection{Testcases and enablements}
\label{subsec:testcases}

To enable studies that are relevant to the sub-10nm regime \cite{GoldieMYJSWLJPNPTSHTLLHCD24}, 
we develop scaled versions of Google's TSMC 7nm Ariane 
testcase ({\em CT-Ariane}), and port other macro-heavy testcases to
the open 7nm enablement ASAP7 \cite{ASAP7}.\footnote{We drop 
the ICCAD04~\cite{iccad04} testcases,
which correspond to much older technologies. Experimental results for 
these testcases  remain available in \cite{TILOS-repo}.} 
Details of testcases and design enablements used in our studies
are as follows.
  
\noindent  
{\bf Testcases.}
We convert the only publicly available Google testcase, Ariane
in TSMC 7nm ({\em CT-Ariane}), from protobuf format (available in the 
{\em CT} repo \cite{CT-TCAD}) to LEF/DEF format~\cite{rdf2024}. 
Scaled (x2 and x4) versions of {\em CT-Ariane} serve as additional
testcases, and are also used in
pre-training and ``quantified suboptimality'' analyses~\cite{HagenHK95}. 
Table \ref{tab:testcase_pretraining} provides details of scaled versions
of {\em CT-Ariane}.\footnote{Note that the CT-Grouping flow uses hMETIS 
to generate standard-cell clusters with a parameter {\em npart}, which
is set to 500 plus the number of predefined groups (macros and IO ports).
Therefore, the  number of standard-cell clusters (\#Grps) scales sublinearly 
in Table \ref{tab:testcase_pretraining}.}
% We do not use it in our study.}

\begin{table}[]
\caption{
%Testcases for pre-training.
Google's TSMC 7nm Ariane testcase ({\em CT-Ariane}) and its scaled versions.
\#Grps indicates the number of standard-cell clusters.}
%and \#Macros represent the number of standard-cell clusters and macros, respectively.}
\label{tab:testcase_pretraining}
\vspace{-0.05in}
\centering
\resizebox{0.9\columnwidth}{!}{%
\begin{tabular}{|c|c|c|c|c|}
\hline
\textbf{Design} & \textbf{\#StdCells (K)} & \textbf{\#Grps} 
        & \textbf{\#Macros}
                & \textbf{\#MacroType} \\ \hline
CT-Ariane       & 83 & 799    & 133
                & 1 \\ \hline
CT-Ariane-X2    & 166 & 982    &  266
                & 1 \\ \hline
CT-Ariane-X4    & 332 & 1519   & 532
                & 1 \\ \hline
\end{tabular}%
}
\end{table}

We also study three open-source testcases which are
publicly available in \cite{MacroPlacementTestcases}: 
Ariane \cite{ariane}, BlackParrot (Quad-Core) \cite{bp_quad} 
and MemPoolGroup \cite{mempool_group}.
%We also study three open-source testcases available in  {\em MacroPlacement}~\cite{MacroPlacementTestcases} includes  Ariane \cite{ariane}, BlackParrot (Quad-Core) \cite{bp_quad} and  MemPool Group \cite{mempool_group}.\footnote{{\em MacroPlacement}~\cite{MacroPlacementTestcases} also includes NVDLA (partition ``c'') \cite{nvdla} testcase.}  and NVDLA (partition ``c'') \cite{nvdla}.
Table \ref{tab:testcase} provides testcase parameters.
\#MacroType gives the number of distinct macro sizes: Ariane has all same-sized
macros, while BlackParrot and MemPoolGroup each contain macros of varying sizes.
We use the 133-macro Ariane variant to match the Ariane in {\em Nature} and {\em CT}.
Table \ref{tab:testcase} also gives ranges for standard cell counts, since \#StdCells differs
between NanGate45 and ASAP7.

\begin{table}[]
\caption{Testcases for evaluation. 
\#FFs and \#Macros respectively represent 
the number of flip-flops and macros in both NanGate45 and ASAP7.}
\label{tab:testcase}
\vspace{-0.05in}
\centering
%\resizebox{0.9\columnwidth}{!}{%
\begin{tabular}{|c|c|c|c|c|}
\hline
\textbf{Design} &\textbf{\#StdCells (K)} & \textbf{\#FFs (K)} & \textbf{\#Macros}
                & \textbf{\#MacroType} \\ \hline
Ariane          & 99 - 117 & 20    & 133
                & 1 \\ \hline
% NVDLA           & 45    & 128
%                 & 1 \\ \hline
BlackParrot     & 686 - 835 & 214   & 220
                & 6 \\ \hline
MemPoolGroup    & 2529 - 2729 & 361   & 324
                & 4 \\ \hline
\end{tabular}%
%}
\end{table}

\noindent
{\bf Enablements.}
Our studies use two open-source enablements that are public in 
{\em MacroPlacement}:
NanGate45 \cite{nangate45} and  ASAP7 \cite{ASAP7}. 
We use the bsg\_fakeram \cite{bsg_fakeram} generator to 
generate SRAMs for NanGate45. 
%The SKY130HD PDK has only five metal layers, while SRAMs  typically use or block the first four metal layers; this makes it difficult to route macro-heavy testcases. We therefore provide the SKY130HD FakeStack~\cite{SKY130HDFakeStack}  enablement which contains nine metal layers. 
We also use FakeRAM2.0~\cite{fakeram2} to generate SRAM
abstracts for ASAP7-based testcases. 
We also use one closed-source enablement:
GlobalFoundries 12LP with SRAMs from a third-party IP provider.
%For our experiments we mainly focus on NanGate45~\cite{nangate45} and ASAP7~\cite{ASAP7}.

\subsection{Commercial evaluation flow}
\label{subsec:evalflow}
\begin{figure}[!htb]
    \centering
    \includegraphics[ width=0.90\columnwidth]{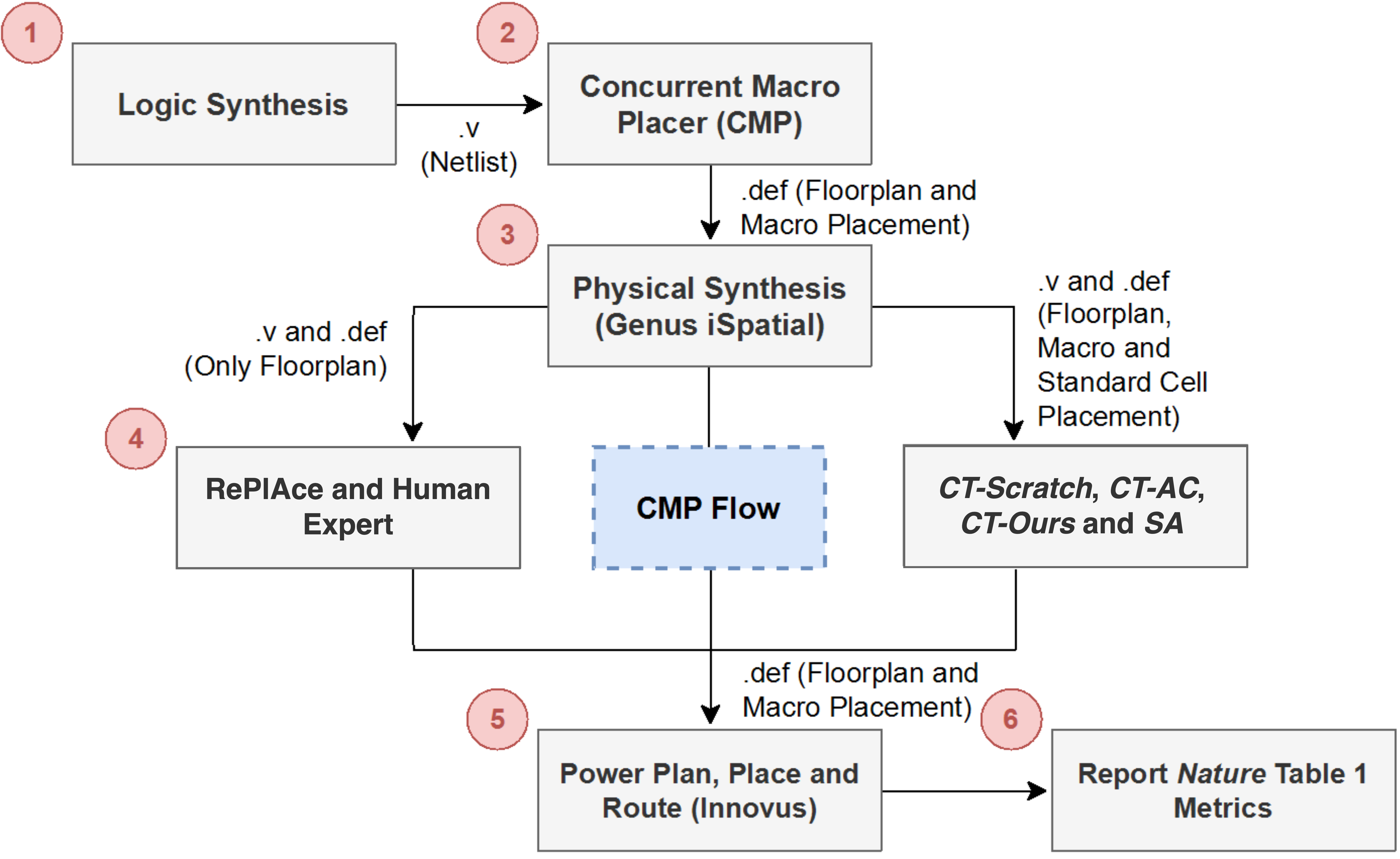}
    \caption{Evaluation flow for placements 
    produced by different macro placers.}
    \label{fig:evalflow}
    \vspace{-0.05in}
\end{figure}

Figure~\ref{fig:evalflow} presents the commercial tool-based
flow that we use to create macro placement instances and 
evaluate macro placement solutions.\footnote{We do not perform
any benchmarking of the EDA tools used in this study.}
The flow has the following steps.

\noindent
\textbf{Step 1:} We run logic synthesis using Cadence Genus 21.1 to 
synthesize a gate-level netlist for a given testcase.

\noindent
\textbf{Step 2:} We input the synthesized netlist to Cadence
Innovus 21.1 and use CMP (Concurrent Macro Placer) to place macros.

\noindent
\textbf{Step 3:} We input the floorplan .def with placed macros 
to the Cadence Genus iSpatial flow and run physical-aware
synthesis, which also generates initial placement locations 
(i.e., (x,y) coordinates) for all standard cells.

\noindent
\textbf{Step 4:} We obtain macro placement solutions from seven
methods: {\em CT-Scratch}, {\em CT-AC}, {\em CT-Ours}, {\em SA}, 
RePlAce, 
%AutoDMP, 
CMP
and human-expert. The CMP macro placement 
is produced in Step 2.
Before running {\em CT} or {\em SA} macro placement,
we convert the Verilog netlist to protocol buffer 
(protobuf) format using code available in \cite{TILOS-repo},
and use CT-Grouping to generate standard-cell 
clusters. The initial placement of standard cells obtained in 
Step 3 is used to guide the 
CT-Grouping process~\cite{GoldieMYJSWLJPNPTSHTLLHCD24}. 
%Code to generate the protobuf netlist and group standard  cells are available in  {\em MacroPlacement}.
For the {\em CT} and {\em SA} runs reported below, we run the 
grouping flow provided in Google's {\em CT} after 
generating the protobuf netlist.
% To generate a Bookshelf-format input netlist for RePlAce, we 
% use the LEF/DEF to Bookshelf converter from 
% RosettaStone \cite{RosettaStone}.
RePlAce and human experts are not given any
initial placement information for standard cells or macros.

\noindent
\textbf{Step 5:} For each macro placement solution, 
we input the floorplan .def with macro placement locations
to Innovus for place and route (P\&R). 
After reading the .def file into Innovus, we set all 
standard cells to unplaced, and legalize macro locations 
using the {\em refine\_macro\_placement}
command.\footnote{Macro placements produced
by RePlAce, {\em CT-Scratch}, {\em CT-AC}, {\em CT-Ours}
and {\em SA} can have macros that are not 
placed on grids (cf. Subsection \ref{subsec:mismatch}).}
We then perform power delivery network (PDN)
generation.\footnote{{\em CT} assumes that 18\% of 
routing tracks are used by PDN \cite{CTPDN}. We implement 
our PDN scripts following the ``18\% rule'' for all 
the enablements. All of our PDN scripts are available at 
\cite{MPPDN} in {\em MacroPlacement}.}
After PDN generation, we run placement, clock tree 
synthesis, routing and post-route optimization (postRouteOpt).

\noindent
\textbf{Step 6:} We extract the total routed wirelength (rWL), standard cell area, 
total power, worst negative slack (WNS), total negative 
slack (TNS) and DRC count from the post-routed design. 
Table 1 of the {\em Nature} paper \cite{MirhoseiniGYJSWLJPNPTSHTLLHCD21}
presents similar metrics to compare different macro placement 
solutions.\footnote{According to {\em Nature} authors,
``The final metrics in Table 1 are reported after PlaceOpt,
meaning that global routing has been performed by the EDA tool''. 
In this paper, we report metrics after postRouteOpt, meaning 
that the entire P\&R flow has been performed. A number of
results reported in \cite{OurProgress} include metrics
after both PlaceOpt and postRouteOpt.} Below, we refer to 
these as the ({\em Nature}) ``{\em Table 1} metrics''.

% \subsection{CT and CT Pretraining details}
% \label{usbsec:pretrain}

\subsection{Settings for Circuit Training}
\label{subsec:CT_Training}

When running Circuit Training, we follow the default setting
given in \cite{CT-TCAD}, except that we 
use a density weight of 0.5 instead of 1.0, based on 
guidance from Google engineers \cite{GWu22}. Training is conducted 
on a single server equipped with eight NVIDIA V100 GPUs, paired 
with five collect servers, each featuring a 96-thread CPU. 
Each training job deploys 256 collect jobs, as increasing 
beyond this threshold yields diminishing speedup benefits~\cite{MirhoseiniGYJSWLJPNPTSHTLLHCD21}. 
Unless otherwise mentioned, we use 333 as the global seed. 
As detailed in Subsection \ref{subsec:compute_resource}, 
we use 200 iterations~\cite{CT-Iteration} for training 
and 400 iterations for fine-tuning across all testcases.
An additional 200 iterations are allocated (i.e., twice the
suggested iterations in the {\em CT}
repo~\cite{Ariane-pre-train}) to ensure that
{\em CT} has sufficient opportunity to achieve its best
results. Due to significant runtime and resource usage,
we cap the extension at 200 iterations.
We then report the best placement cost found across all iterations. 
For training that diverges, we restart the run at least two 
more times to confirm the divergence.\footnote{One exception: training
of {\em CT-Ariane-X4} from scratch was abandoned after loss and placement return stayed flat in
three attempts with 200 iterations.}

\section{Experiments and Results}
\label{sec:experiment}

In this section, we first evaluate the performance of {\em CT} and other macro placers.
We then present results of stability and ablation studies. All scripts are public
in \cite{TILOS-repo}.

\subsection{Comparison of CT with other macro placers}
\label{subsec:comparison}

\noindent
{\bf Configuration of different macro placers.} We generate 
macro placement solutions using {\em CT}, CMP, {\em SA} and RePlAce.
We also include macro placement solutions generated 
by human-experts. 
For {\em CT}, we provide three results: training AlphaChip (i.e., the latest version 
of the renamed framework in the Circuit Training repository \cite{CT-TCAD}) from scratch 
(denoted as {\em CT-Scratch}); fine-tuning AlphaChip using the pre-trained checkpoint released 
in August 2024 (denoted as {\em CT-AC});
and fine-tuning AlphaChip using our own checkpoint pre-trained with specific
testcase variants (denoted as {\em CT-Ours} (see Subsection \ref{subsec:pre-train-slice} for details).
Due to the very high runtime of pre-training, we
run {\em CT-Ours} only for the Ariane designs.
We follow the default settings in \cite{CT-TCAD},
except that we use a density weight of 0.5 instead of 1.0, based on guidance
from Google engineers \cite{GWu22}.
For CMP, we use the default tool settings. 
For {\em SA}, we use the configurations described in
Section~\ref{sec:sa} and the C++ implementation available in \cite{TILOS-repo}.
For RePlAce, we use OpenROAD~\cite{RePlAce} and set the target density as
$util + (1-util)*0.5$, where $util$ is the floorplan density of the design.
All experiments use Genus 21.1 for synthesis (Fig. \ref{fig:evalflow}, Step 1)
and Innovus 21.1 for P\&R (Steps 5 and 6).

\noindent
{\bf Evaluation of {\em Nature} {\em Table 1} metrics for different macro placers.}
We generate macro placement solutions for testcases in open NanGate45
(NG45) and ASAP7 and commercial GlobalFoundries 12nm (GF12) enablements.
Table \ref{tab:mp_result} presents {\em Nature} 
{\em Table 1} metrics obtained using the evaluation flow of 
Figure \ref{fig:evalflow} for different macro placers on our
testcases. The {\em Table 1} metrics in GF12 are normalized to protect foundry IP: 
(i) standard-cell area is normalized to core area; (ii) total 
power and routed wirelength (rWL) are normalized to the {\em CT-AC} result; 
and (iii) timing metrics (WNS, TNS) are normalized to the target clock 
period (TCP) which we leave unspecified.\footnote{WNS and TNS
timing metrics in Table~\ref{tab:mp_result} suggest that TCP for MemPoolGroup-GF12 could
be increased; we report such improvements and updates 
in \cite{OurProgress}.}
% XXX ABK -- question about normalization in Table III ...
In NG45, the default TCP values for Ariane, BlackParrot Quad-Core
(BlackParrot), and MemPoolGroup (MemPool) are 1.3ns, 1.3ns, and 4.0ns.
In ASAP7, they are 0.9ns, 0.85ns, and 1.8ns.
All testcases reported in Table \ref{tab:mp_result} have 68\%
floorplan utilization, matching the Ariane design that is public in {\em CT}.

\begin{figure*}
    \centering
    \includegraphics[width=1.9\columnwidth]{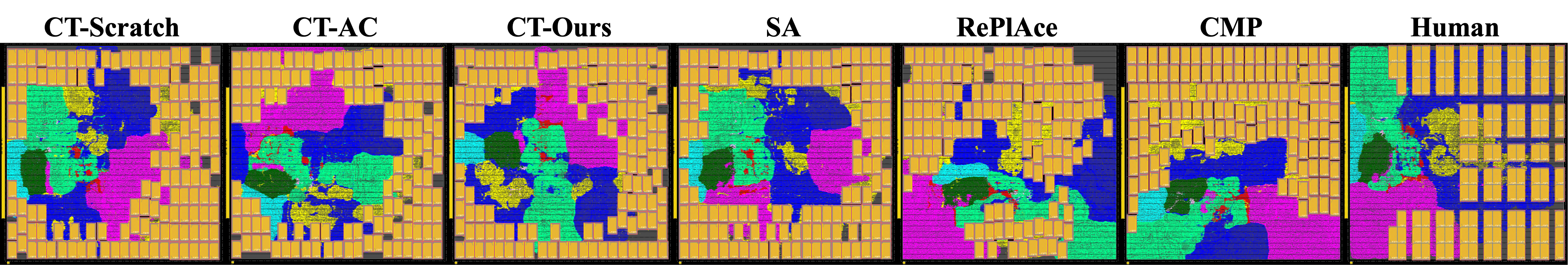}
    \vspace{-0.1in}
    \centering
    \caption{Macro placement solutions for the Ariane-ASAP7 design, produced by different macro placers.}
    \label{fig:ariane_macroplacement}
    \vspace{-0.2in}
\end{figure*}

Table \ref{tab:mp_result} also reports
{\em CT} proxy cost for all macro placement solutions, as evaluated by {\em CT}'s
{\em plc\_client}. To compute proxy cost for 
CMP, RePlAce, {\em SA} and human-expert solutions, we first 
update hard macro locations and orientations, then run FD placement
via the {\em plc\_client} to place all standard-cell clusters
(soft macros). We then compute the proxy cost. Figure~\ref{fig:ariane_macroplacement}
shows Ariane-ASAP7 macro placements
produced by the macro placers we study.\footnote{\label{fn:drc}\sk{All
postRouteOpt designs are DRC-clean, except that in our ASAP7 enablement,
we observe spacing violations across all macro placement solutions around
macro blockages for MemPool Group design. These violations arise from a
few macro pins being covered by the blockage layer, resulting in total
DRC counts below 200.}}
We observe the following.
\begin{itemize}[noitemsep, topsep=0pt, leftmargin=*]
    \item {\bf Comparison of routed wirelength (rWL):}
    CMP consistently dominates the other macro placers except on Ariane-NG45,
    where CMP's rWL is 2\% worse than {\em SA}'s.
    For BlackParrot-GF12, CMP has 34\% less rWL than {\em CT-AC}.

    \item {\bf Comparison of proxy cost:} {\em SA} dominates other macro placers 
    in 6 of 9 cases, {\em CT-AC} dominates other macro placers in 2 of 9 cases, 
    and {\em CT-Scratch} dominates other macro placers in 1 of 9 cases.

    \item {\bf Comparison between {\em CT-AC} and {\em CT-Ours}:}
    {\em CT-AC} consistently outperforms {\em CT-Ours} in proxy cost for the 
    Ariane testcase in all three technologies (NG45, ASAP7, GF12).

    \item {\bf Comparison between {\em CT-AC} and {\em SA}:}
    {\em CT-AC} yields better TNS than {\em SA} for 6 of 9 cases,
    while {\em SA} gives better rWL (7 of 9 cases) and proxy cost (6 of 9 cases) than {\em CT-AC}.
    
    \item {\bf Comparison between {\em CT-AC} and Human experts:}
    For large macro-heavy designs such as BlackParrot and MemPoolGroup, 
    human experts dominate {\em CT-AC} in terms of the {\em Nature Table 1}
    metrics of postRouteOpt ``ground truth''~\cite{MirhoseiniGYJSWLJPNPTSHTLLHCD21}
    outcomes (5 of 6 cases). For these large designs,
    even though the proxy congestion cost for {\em CT-AC}
    is better than that of the human expert, both macro
    placement solutions finish routing DRC-clean or with
    very similar DRCs in ASAP7, as mentioned in
    Footnote~\ref{fn:drc}.

    \item {\bf Comparison of resource usage:} Resource requirements vary across
    the macro placers. Considering that 1 GPU is equivalent to 10 CPUs, i.e., following the analysis 
    in~\cite{MirhoseiniGYJSWLJPNPTSHTLLHCD21},
    resources ($\#cpus \times runtime$) used for BlackParrot-NG45 by
    {\em CT-Scratch}, {\em CT-AC}, {\em SA}, RePlAce, and CMP are, respectively,
    $(8\times10 + 256)\times56.76 = 19,068$, $(8\times10 + 256)\times64.01
    = 21,507$, $80\times11.2 = 896$, $1\times0.31 = 0.31$, and
    $8\times0.33 = 2.64$ CPU-hours.\footnote{Or, in financial terms, running the {\em CT} model 
    on Google Cloud~\cite{gcpprice} with one 8 NVIDIA V100 GPU train server and 
    five collect servers costs approximately $\$(17.01 + 5\times3.21)=\$33.06$/hr, while {\em SA} runs 
    cost $\$3.21$/hr.} Human expert solutions are generated in under 12 hours for each design.
    \item {\bf {\em SA} runtime calibration:} We report {\em SA} runtimes on
    an Intel Xeon Gold 6148 CPU. However, we have observed that for some testcases,
    runtime can be as low as one-third of this reported runtime 
    when using an AMD EPYC 7742 CPU.
    \item {\bf Comparison of {\em SA} with {\em SA} in~\cite{ispd23}:} 
    Compared to the {\em SA} in~\cite{ispd23}, our updated {\em SA} with GWTW
    achieves better proxy cost in two of the three NG45 testcases 
    (geometric-mean improvement of 9\% and up to 26\%), better routed wirelength
    and total power in all three cases, and better TNS in one of the three
    cases. \sk{Note that {\em SA} in~\cite{ispd23} finds
    better proxy values than {\em CT-AC} for two of the three NG45 testcases.}
\end{itemize}

\begin{table*}[]
\caption{PPA, proxy cost, runtime, and resource details of
different macro placement solutions on our modern benchmarks,
across three different enablements. GF12-based results are
normalized. $^\dagger$ indicates that a total of 400 iterations are 
used for training. Best rWL, TNS and proxy cost values for each design,
across all macro placement methods, are highlighted using blue bold font. 
\sk{\#G and \#C denote the number of V100 GPUs and CPU threads, respectively.}}
% \red{Ensure all bold blue fonts are correctly highlighted.}
\label{tab:mp_result}
% \sk{@YW, please add the CT runtime information.}

\centering
% \resizebox{1.95\columnwidth}{!}{%
\begin{tabular}{|c|c|ccccr|cccc|c|}
\hline
\multirow{2}{*}{\begin{tabular}[c]{@{}c@{}}Design \\ Tech\end{tabular}} &
  \multirow{2}{*}{\begin{tabular}[c]{@{}c@{}}Macro \\ Placer\end{tabular}} &
  \multicolumn{5}{c|}{PostRouteOpt PPA  (From Innovus)} &
  \multicolumn{4}{c|}{Proxy Cost Details} &
  \multirow{2}{*}{\begin{tabular}[c]{@{}c@{}}Runtime (Hrs) \\ (\#G, \#C)\end{tabular}} \\ \cline{3-11}
 &
   &
  \multicolumn{1}{c|}{Area ($\mu m^2$)} &
  \multicolumn{1}{c|}{rWL ($\mu m$)} &
  \multicolumn{1}{c|}{Power ($mW$)} &
  \multicolumn{1}{c|}{WNS ($ps$)} &
  TNS ($ns$) &
  \multicolumn{1}{c|}{WL} &
  \multicolumn{1}{c|}{Den.} &
  \multicolumn{1}{c|}{Cong.} &
  Proxy &
   \\ \hline \hline
\multirow{7}{*}{\begin{tabular}[c]{@{}c@{}}Ariane\\ NG45\end{tabular}} &
  CT-Scratch$^\dagger$ &
  \multicolumn{1}{r|}{246303} &
  \multicolumn{1}{r|}{4648156} &
  \multicolumn{1}{r|}{832} &
  \multicolumn{1}{r|}{-140} &
  -119.1 &
  \multicolumn{1}{c|}{0.102} &
  \multicolumn{1}{c|}{0.518} &
  \multicolumn{1}{c|}{0.973} &
  0.848 &
  36.26 (8, 576) \\ \cline{2-12} 
 &
  CT-AC$^\dagger$ &
  \multicolumn{1}{r|}{248382} &
  \multicolumn{1}{r|}{4995968} &
  \multicolumn{1}{r|}{836} &
  \multicolumn{1}{r|}{-88} &
  -52.2 &
  \multicolumn{1}{c|}{0.101} &
  \multicolumn{1}{c|}{\bf \blue{0.508}} &
  \multicolumn{1}{c|}{0.931} &
  0.820 &
  36.18 (8, 576) \\ \cline{2-12} 
 &
  CT-Ours$^\dagger$ &
  \multicolumn{1}{r|}{245703} &
  \multicolumn{1}{r|}{4898125} &
  \multicolumn{1}{r|}{831} &
  \multicolumn{1}{r|}{-86} &
  -57.8 &
  \multicolumn{1}{c|}{0.108} &
  \multicolumn{1}{c|}{0.538} &
  \multicolumn{1}{c|}{0.966} &
  0.860 &
  38.79 (8, 576) \\ \cline{2-12} 
 &
  {\em SA} &
  \multicolumn{1}{r|}{247777} &
  \multicolumn{1}{r|}{\bf \blue{3976569}} &
  \multicolumn{1}{r|}{827} &
  \multicolumn{1}{r|}{-126} &
  -116.8 &
  \multicolumn{1}{c|}{0.090} &
  \multicolumn{1}{c|}{0.515} &
  \multicolumn{1}{c|}{\bf \blue{0.907}} &
  {\bf \blue{0.801}} &
  11.52 (0, 80) \\ \cline{2-12} 
 &
  RePlAce &
  \multicolumn{1}{r|}{251117} &
  \multicolumn{1}{r|}{5131963} &
  \multicolumn{1}{r|}{842} &
  \multicolumn{1}{r|}{-99} &
  -94.0 &
  \multicolumn{1}{c|}{0.092} &
  \multicolumn{1}{c|}{0.998} &
  \multicolumn{1}{c|}{1.748} &
  1.465 & 0.04 (0, 1)
   \\ \cline{2-12} 
 &
  CMP &
  \multicolumn{1}{r|}{256230} &
  \multicolumn{1}{r|}{4057140} &
  \multicolumn{1}{r|}{852} &
  \multicolumn{1}{r|}{-154} &
  -196.5 &
  \multicolumn{1}{c|}{\bf \blue{0.088}} &
  \multicolumn{1}{c|}{0.909} &
  \multicolumn{1}{c|}{1.455} &
  1.270 & 0.05 (0, 8)
   \\ \cline{2-12}
 &
  Human &
  \multicolumn{1}{r|}{249034} &
  \multicolumn{1}{r|}{4681178} &
  \multicolumn{1}{r|}{832} &
  \multicolumn{1}{r|}{-88} &
  {\bf \blue{-46.8}} &
  \multicolumn{1}{c|}{0.107} &
  \multicolumn{1}{c|}{0.738} &
  \multicolumn{1}{c|}{1.376} &
  1.164 & NA
   \\ \hline \hline
\multirow{6}{*}{\begin{tabular}[c]{@{}c@{}}BlackParrot\\ NG45\end{tabular}} &
  CT-Scratch$^\dagger$ &
  \multicolumn{1}{r|}{1990312} &
  \multicolumn{1}{r|}{47785761} &
  \multicolumn{1}{r|}{4822} &
  \multicolumn{1}{r|}{-205} &
  -1203.4 &
  \multicolumn{1}{c|}{0.096} &
  \multicolumn{1}{c|}{0.790} &
  \multicolumn{1}{c|}{1.132} &
  1.057 &
  56.76 (8, 576) \\ \cline{2-12} 
 &
  CT-AC$^\dagger$ &
  \multicolumn{1}{r|}{1944272} &
  \multicolumn{1}{r|}{33165774} &
  \multicolumn{1}{r|}{4569} &
  \multicolumn{1}{r|}{-230} &
  -1486.5 &
  \multicolumn{1}{c|}{0.066} &
  \multicolumn{1}{c|}{0.755} &
  \multicolumn{1}{c|}{1.053} &
  0.970 &
  64.01 (8, 576) \\ \cline{2-12} 
 
 &
  {\em SA} &
  \multicolumn{1}{r|}{1938779} &
  \multicolumn{1}{r|}{28937792} &
  \multicolumn{1}{r|}{4512} &
  \multicolumn{1}{r|}{-230} &
  -3012.8 &
  \multicolumn{1}{c|}{0.054} &
  \multicolumn{1}{c|}{\bf \blue{0.711}} &
  \multicolumn{1}{c|}{\bf \blue{0.936}} &
  {\bf \blue{0.878}} &
  11.20 (0, 80) \\ \cline{2-12} 
 &
  RePlAce &
  
  \multicolumn{1}{r|}{1930960} &
  \multicolumn{1}{r|}{26854143} &
  \multicolumn{1}{r|}{4485} &
  \multicolumn{1}{r|}{-191} &
  -868.0 &
  \multicolumn{1}{c|}{0.050} &
  \multicolumn{1}{c|}{1.049} &
  \multicolumn{1}{c|}{1.153} &
  1.151 & 0.31 (0, 1)
   \\ \cline{2-12} 
 &
  CMP &
  \multicolumn{1}{r|}{1916166} &
  \multicolumn{1}{r|}{\bf \blue{23144317}} &
  \multicolumn{1}{r|}{4429} &
  \multicolumn{1}{r|}{-144} &
  -356.2 &
  \multicolumn{1}{r|}{\bf \blue{0.050}} &
  \multicolumn{1}{r|}{0.882} &
  \multicolumn{1}{r|}{1.066} &
  1.024 & 0.33 (0, 8)
   \\ \cline{2-12} 
 &
  Human &
  \multicolumn{1}{r|}{1919928} &
  \multicolumn{1}{r|}{25915520} &
  \multicolumn{1}{r|}{4470} &
  \multicolumn{1}{r|}{-97} &
  \bf \blue{-321.9} &
  \multicolumn{1}{c|}{0.054} &
  \multicolumn{1}{c|}{1.158} &
  \multicolumn{1}{c|}{1.260} &
  1.263 & NA
   \\ \hline \hline
\multirow{6}{*}{\begin{tabular}[c]{@{}c@{}}MemPool\\ 
  Group\\ NG45\end{tabular}} &
  CT-Scratch$^\dagger$ &
  \multicolumn{1}{r|}{4915555} &
  \multicolumn{1}{r|}{119607588} &
  \multicolumn{1}{r|}{2754} &
  \multicolumn{1}{r|}{-163} &
  -47.7 &
  \multicolumn{1}{c|}{0.064} &
  \multicolumn{1}{c|}{1.200} &
  \multicolumn{1}{c|}{1.232} &
  1.280 &
  91.77 (8, 576) \\ \cline{2-12} 
 &
  CT-AC$^\dagger$ &
  \multicolumn{1}{r|}{4871665} &
  \multicolumn{1}{r|}{112486298} &
  \multicolumn{1}{r|}{2683} &
  \multicolumn{1}{r|}{-51} &
  -32.1 &
  \multicolumn{1}{c|}{0.062} &
  \multicolumn{1}{c|}{\blue{\bf 1.006}} &
  \multicolumn{1}{c|}{\blue{\bf 1.086}} &
  \blue{\bf 1.108} &
  91.39 (8, 576) \\ \cline{2-12} 
 
 &
  {\em SA} &
  \multicolumn{1}{r|}{4915598} &
  \multicolumn{1}{r|}{115229509} &
  \multicolumn{1}{r|}{2720} &
  \multicolumn{1}{r|}{-32} &
  -5.9 &
  \multicolumn{1}{c|}{0.062} &
  \multicolumn{1}{c|}{1.131} &
  \multicolumn{1}{c|}{1.095} &
  1.175 &
  11.90 (0, 80) \\ \cline{2-12} 
 &
  RePlAce &
  \multicolumn{1}{r|}{4930394} &
  \multicolumn{1}{r|}{113315081} &
  \multicolumn{1}{r|}{2688} &
  \multicolumn{1}{r|}{-96} &
  -7.3 &
  \multicolumn{1}{c|}{\bf \blue{0.056}} &
  \multicolumn{1}{c|}{1.621} &
  \multicolumn{1}{c|}{1.652} &
  1.693 & 0.88 (0, 1)
   \\ \cline{2-12} 
 &
  CMP &
  \multicolumn{1}{r|}{4837150} &
  \multicolumn{1}{r|}{\bf \blue{102907484}} &
  \multicolumn{1}{r|}{2587} &
  \multicolumn{1}{r|}{-20} &
  {\bf \blue{-1.0}} &
  \multicolumn{1}{c|}{0.057} &
  \multicolumn{1}{c|}{1.495} &
  \multicolumn{1}{c|}{1.554} &
  1.581 & 1.97 (0, 8)
   \\ \cline{2-12}
 &
  Human &
  \multicolumn{1}{r|}{4873872} &
  \multicolumn{1}{r|}{107597894} &
  \multicolumn{1}{r|}{2640} &
  \multicolumn{1}{r|}{-49} &
  -11.9 &
  \multicolumn{1}{c|}{0.067} &
  \multicolumn{1}{c|}{1.586} &
  \multicolumn{1}{c|}{1.710} &
  1.715 & NA
   \\ \hline \hline
\multirow{7}{*}{\begin{tabular}[c]{@{}c@{}}Ariane\\ ASAP7\end{tabular}} &
  CT-Scratch$^\dagger$ &
  \multicolumn{1}{r|}{16570} &
  \multicolumn{1}{r|}{1026239} &
  \multicolumn{1}{r|}{505} &
  \multicolumn{1}{r|}{-142} &
  -184.2 &
  \multicolumn{1}{c|}{0.119} &
  \multicolumn{1}{c|}{0.821} &
  \multicolumn{1}{c|}{0.871} &
  0.965 &
  36.53 (8, 576) \\ \cline{2-12} 
 &
  
  CT-AC &
  \multicolumn{1}{r|}{16524} &
  \multicolumn{1}{r|}{1014938} &
  \multicolumn{1}{r|}{505} &
  \multicolumn{1}{r|}{-108} &
  -105.0 &
  \multicolumn{1}{c|}{0.122} &
  \multicolumn{1}{c|}{0.804} &
  \multicolumn{1}{c|}{0.850} &
  0.950 &
  36.35 (8, 576) \\ \cline{2-12}
 &
  CT-Ours$^\dagger$ &
  \multicolumn{1}{r|}{16612} &
  \multicolumn{1}{r|}{1033863} &
  \multicolumn{1}{r|}{505} &
  \multicolumn{1}{r|}{-144} &
  -204.1 &
  \multicolumn{1}{c|}{0.125} &
  \multicolumn{1}{c|}{\bf \blue{0.811}} &
  \multicolumn{1}{c|}{0.873} &
  0.967 &
  39.95 (8, 576) \\ \cline{2-12} 
 &
  {\em SA} &
  \multicolumn{1}{r|}{16467} &
  \multicolumn{1}{r|}{886776} &
  \multicolumn{1}{r|}{503} &
  \multicolumn{1}{r|}{-124} &
  -141.1 &
  \multicolumn{1}{c|}{0.108} &
  \multicolumn{1}{c|}{0.817} &
  \multicolumn{1}{c|}{\bf \blue{0.822}} &
  {\bf \blue{0.928}} &
  10.87 (0, 80) \\ \cline{2-12} 
 &
  RePlAce &
  \multicolumn{1}{r|}{16410} &
  \multicolumn{1}{r|}{917539} &
  \multicolumn{1}{r|}{504} &
  \multicolumn{1}{r|}{-108} &
  -124.0 &
  \multicolumn{1}{c|}{\bf \blue{0.102}} &
  \multicolumn{1}{c|}{1.169} &
  \multicolumn{1}{c|}{1.160} &
  1.266 & 0.02 (0, 1)
   \\ \cline{2-12} 
 &
  CMP &
  \multicolumn{1}{r|}{16350} &
  \multicolumn{1}{r|}{\bf \blue{843757}} &
  \multicolumn{1}{r|}{504} &
  \multicolumn{1}{r|}{-124} &
  -146.1 &
  \multicolumn{1}{c|}{0.102} &
  \multicolumn{1}{c|}{1.122} &
  \multicolumn{1}{c|}{1.141} &
  1.233 & 0.04 (0, 8)
   \\ \cline{2-12}
 &
  Human &
  \multicolumn{1}{r|}{16613} &
  \multicolumn{1}{r|}{1182350} &
  \multicolumn{1}{r|}{508} &
  \multicolumn{1}{r|}{-104} &
  {\bf \blue{-81.8}} &
  \multicolumn{1}{c|}{0.131} &
  \multicolumn{1}{c|}{1.177} &
  \multicolumn{1}{c|}{1.484} &
  1.461 & NA
   \\ \hline \hline
\multirow{6}{*}{\begin{tabular}[c]{@{}c@{}}BlackParrot\\ ASAP7\end{tabular}} &
  CT-Scratch$^\dagger$ &
  \multicolumn{1}{r|}{126524} &
  \multicolumn{1}{r|}{11380551} &
  \multicolumn{1}{r|}{1609} &
  \multicolumn{1}{r|}{-226} &
  -2043.7 &
  \multicolumn{1}{c|}{0.089} &
  \multicolumn{1}{c|}{0.908} &
  \multicolumn{1}{c|}{1.002} &
  1.044 &
  55.87 (8, 576) \\ \cline{2-12} 
 &
  CT-AC$^\dagger$ &
  \multicolumn{1}{r|}{124987} &
  \multicolumn{1}{r|}{8880315} &
  \multicolumn{1}{r|}{1569} &
  \multicolumn{1}{r|}{-201} &
  -1448.6 &
  \multicolumn{1}{c|}{0.067} &
  \multicolumn{1}{c|}{0.848} &
  \multicolumn{1}{c|}{0.833} &
  0.908 &
  58.27 (8, 576) \\ \cline{2-12} 
 
 &
  {\em SA} &
  \multicolumn{1}{r|}{123141} &
  \multicolumn{1}{r|}{7266869} &
  \multicolumn{1}{r|}{1547} &
  \multicolumn{1}{r|}{-120} &
  -424.8 &
  \multicolumn{1}{c|}{\bf \blue{0.053}} &
  \multicolumn{1}{c|}{\bf \blue{0.758}} &
  \multicolumn{1}{c|}{\bf \blue{0.751}} &
  {\bf \blue{0.808}} &
  9.78 (0, 80) \\ \cline{2-12} 
 &
  RePlAce &
  \multicolumn{1}{r|}{123205} &
  \multicolumn{1}{r|}{6718623} &
  \multicolumn{1}{r|}{1540} &
  \multicolumn{1}{r|}{-96} &
  -590.0 &
  \multicolumn{1}{c|}{0.064} &
  \multicolumn{1}{c|}{1.097} &
  \multicolumn{1}{c|}{1.066} &
  1.145 & 0.20 (0, 1)
   \\ \cline{2-12} 
 &
  CMP &
  \multicolumn{1}{r|}{122603} &
  \multicolumn{1}{r|}{\bf \blue{6104230}} &
  \multicolumn{1}{r|}{1529} &
  \multicolumn{1}{r|}{-111} &
  {\bf \blue{-240.4}} &
  \multicolumn{1}{c|}{0.058} &
  \multicolumn{1}{c|}{1.058} &
  \multicolumn{1}{c|}{0.936} &
   1.055 & 0.65 (0, 8)
   \\ \cline{2-12} 
 &
  Human &
  \multicolumn{1}{r|}{122914} &
  \multicolumn{1}{r|}{6521501} &
  \multicolumn{1}{r|}{1536} &
  \multicolumn{1}{r|}{-89} &
  -356.6 &
  \multicolumn{1}{c|}{0.057} &
  \multicolumn{1}{c|}{1.204} &
  \multicolumn{1}{c|}{1.053} &
  1.186 & NA
   \\ \hline \hline
\multirow{6}{*}{\begin{tabular}[c]{@{}c@{}}MemPool\\Group\\ ASAP7\end{tabular}} &
  CT-Scratch$^\dagger$ &
  \multicolumn{10}{c|}{Divergence} \\
 \cline{2-12} 
 &
  CT-AC$^\dagger$ &
  \multicolumn{1}{r|}{339535} &
  \multicolumn{1}{r|}{27208664} &
  \multicolumn{1}{r|}{1402} &
  \multicolumn{1}{r|}{-122} &
  -629.0 &
  \multicolumn{1}{c|}{0.072} &
  \multicolumn{1}{c|}{\bf \blue{1.170}} &
  \multicolumn{1}{c|}{\bf \blue{0.812}} &
  {\bf \blue{1.063}} &
  112.87 (8, 576) \\ \cline{2-12} 
 
 &
  {\em SA} &
  \multicolumn{1}{r|}{338798} &
  \multicolumn{1}{r|}{26898162} &
  \multicolumn{1}{r|}{1402} &
  \multicolumn{1}{r|}{-169} &
  -941.0 &
  \multicolumn{1}{c|}{0.069} &
  \multicolumn{1}{c|}{1.305} &
  \multicolumn{1}{c|}{0.834} &
  1.139 &
  10.19 (0, 80) \\ \cline{2-12} 
 &
  RePlAce &
  \multicolumn{1}{r|}{338781} &
  \multicolumn{1}{r|}{26239567} &
  \multicolumn{1}{r|}{1387} &
  \multicolumn{1}{r|}{-152} &
  -819.9 &
  \multicolumn{1}{c|}{0.063} &
  \multicolumn{1}{c|}{1.740} &
  \multicolumn{1}{c|}{1.319} &
  1.593 & 0.60 (0, 1)
   \\ \cline{2-12} 
 &
  CMP &
  \multicolumn{1}{r|}{338559} &
  \multicolumn{1}{r|}{\bf \blue{23259139}} &
  \multicolumn{1}{r|}{1343} &
  \multicolumn{1}{r|}{-88} &
  -224.9 &
  \multicolumn{1}{c|}{\bf \blue{0.060}} &
  \multicolumn{1}{c|}{1.756} &
  \multicolumn{1}{c|}{1.207} &
  1.541 & 1.09 (0, 8)
   \\ \cline{2-12} 
 &
  Human &
  \multicolumn{1}{r|}{338457} &
  \multicolumn{1}{r|}{24573102} &
  \multicolumn{1}{r|}{1354} &
  \multicolumn{1}{r|}{-84} &
  {\bf \blue{-193.4}} &
  \multicolumn{1}{c|}{0.073} &
  \multicolumn{1}{c|}{1.758} &
  \multicolumn{1}{c|}{1.326} &
  1.614 & NA
   \\ \hline \hline
\multirow{7}{*}{\begin{tabular}[c]{@{}c@{}}Ariane\\ GF12\end{tabular}} &
  CT-Scratch &
  \multicolumn{1}{r|}{0.139} &
  \multicolumn{1}{r|}{1.018} &
  \multicolumn{1}{r|}{1.006} &
  \multicolumn{1}{r|}{-0.128} &
  \blue{\bf -106.6} &
  \multicolumn{1}{c|}{0.095} &
  \multicolumn{1}{c|}{0.529} &
  \multicolumn{1}{c|}{0.689} &
  0.704 &
  37.42 (8, 576) \\ \cline{2-12}
 &
 CT-AC &
  \multicolumn{1}{r|}{0.139} &
  \multicolumn{1}{r|}{1.000} &
  \multicolumn{1}{r|}{1.000} &
  \multicolumn{1}{r|}{-0.194} &
  -201.3 &
  \multicolumn{1}{c|}{0.092} &
  \multicolumn{1}{c|}{0.528} &
  \multicolumn{1}{c|}{0.678} &
  0.695 &
  37.22 (8, 576) \\ \cline{2-12}
 &
  CT-Ours$^\dagger$ &
  \multicolumn{1}{r|}{0.139} &
  \multicolumn{1}{r|}{1.093} &
  \multicolumn{1}{r|}{1.016} &
  \multicolumn{1}{r|}{-0.135} &
  -120.9 &
  \multicolumn{1}{c|}{0.107} &
  \multicolumn{1}{c|}{0.554} &
  \multicolumn{1}{c|}{0.705} &
  0.736 &
  35.32 (8, 576) \\ \cline{2-12}
 &
  {\em SA} &
  \multicolumn{1}{r|}{0.138} &
  \multicolumn{1}{r|}{0.894} &
  \multicolumn{1}{r|}{0.993} &
  \multicolumn{1}{r|}{-0.159} &
  -176.2 &
  \multicolumn{1}{c|}{0.092} &
  \multicolumn{1}{c|}{\bf \blue{0.522}} &
  \multicolumn{1}{c|}{\bf \blue{0.677}} &
  {\bf \blue{0.692}} &
  10.16 (0, 80) \\ \cline{2-12} 
 &
  RePlAce &
  \multicolumn{1}{r|}{0.140} &
  \multicolumn{1}{r|}{1.016} &
  \multicolumn{1}{r|}{1.018} &
  \multicolumn{1}{r|}{-0.168} &
  -197.4 &
  \multicolumn{1}{c|}{0.093} &
  \multicolumn{1}{c|}{0.550} &
  \multicolumn{1}{c|}{0.673} &
  0.704 &
  0.03 (0, 1) \\ \cline{2-12}
 &
  CMP &
  \multicolumn{1}{r|}{0.139} &
  \multicolumn{1}{r|}{\blue{\bf 0.843}} &
  \multicolumn{1}{r|}{0.993} &
  \multicolumn{1}{r|}{-0.159} &
  -142.3 &
  \multicolumn{1}{c|}{\bf \blue{0.082}} &
  \multicolumn{1}{c|}{0.748} &
  \multicolumn{1}{c|}{0.831} &
  0.871 &
  0.04 (0, 8) \\ \cline{2-12}
 &
  Human &
  \multicolumn{1}{r|}{0.137} &
  \multicolumn{1}{r|}{1.037} &
  \multicolumn{1}{r|}{0.983} &
  \multicolumn{1}{r|}{-0.139} &
  \blue{\bf -106.6} &
  \multicolumn{1}{c|}{0.104} &
  \multicolumn{1}{c|}{0.914} &
  \multicolumn{1}{c|}{1.156} &
  1.139 &
  NA \\ \hline \hline
\multirow{6}{*}{\begin{tabular}[c]{@{}c@{}}BlackParrot\\ GF12\end{tabular}} &
  CT-Scratch$^\dagger$ &
  \multicolumn{1}{r|}{0.192} &
  \multicolumn{1}{r|}{1.189} &
  \multicolumn{1}{r|}{1.025} &
  \multicolumn{1}{r|}{-0.099} &
  -80.0 &
  \multicolumn{1}{c|}{0.088} &
  \multicolumn{1}{c|}{0.861} &
  \multicolumn{1}{c|}{0.842} &
  0.940 &
  48.10 (8, 576) \\ \cline{2-12} 
 &
  CT-AC$^\dagger$ &
  \multicolumn{1}{r|}{0.191} &
  \multicolumn{1}{r|}{1.000} &
  \multicolumn{1}{r|}{1.000} &
  \multicolumn{1}{r|}{-0.059} &
  -42.7 &
  \multicolumn{1}{c|}{0.087} &
  \multicolumn{1}{c|}{0.754} &
  \multicolumn{1}{c|}{0.752} &
  0.825 &
  51.10 (8, 576) \\ \cline{2-12} 
 
 &
  {\em SA} &
  \multicolumn{1}{r|}{0.190} &
  \multicolumn{1}{r|}{0.867} &
  \multicolumn{1}{r|}{0.978} &
  \multicolumn{1}{r|}{-0.084} &
  -45.5 &
  \multicolumn{1}{c|}{0.058} &
  \multicolumn{1}{c|}{\bf \blue{0.610}} &
  \multicolumn{1}{c|}{\bf \blue{0.640}} &
  {\bf \blue{0.683}} &
  8.90 (0, 80) \\ \cline{2-12} 
 &
  RePlAce &
  \multicolumn{1}{r|}{0.191} &
  \multicolumn{1}{r|}{0.751} &
  \multicolumn{1}{r|}{0.967} &
  \multicolumn{1}{r|}{-0.098} &
  -143.9 &
  \multicolumn{1}{c|}{0.056} &
  \multicolumn{1}{c|}{1.027} &
  \multicolumn{1}{c|}{0.865} &
  1.002 & 0.17 (0, 1)
   \\ \cline{2-12} 
 &
  CMP &
  \multicolumn{1}{r|}{0.190} &
  \multicolumn{1}{r|}{\blue{\bf 0.662}} &
  \multicolumn{1}{r|}{0.949} &
  \multicolumn{1}{r|}{-0.087} &
  -138.9 &
  \multicolumn{1}{c|}{\bf \blue{0.051}} &
  \multicolumn{1}{c|}{0.871} &
  \multicolumn{1}{c|}{0.779} &
  0.876 & 0.29 (0, 8)
   \\ \cline{2-12} 
 &
  Human &
  \multicolumn{1}{r|}{0.189} &
  \multicolumn{1}{r|}{0.709} &
  \multicolumn{1}{r|}{0.954} &
  \multicolumn{1}{r|}{-0.049} &
  \blue{\bf -15.4} &
  \multicolumn{1}{c|}{0.054} &
  \multicolumn{1}{c|}{1.152} &
  \multicolumn{1}{c|}{0.949} &
  1.105 & NA
   \\ \hline \hline
\multirow{6}{*}{\begin{tabular}[c]{@{}c@{}}MemPool\\Group\\ GF12\end{tabular}} &
  CT-Scratch$^\dagger$ &
  \multicolumn{1}{r|}{0.413} &
  \multicolumn{1}{r|}{1.105} &
  \multicolumn{1}{r|}{1.074} &
  \multicolumn{1}{r|}{-0.178} &
  -2201.3 &
  \multicolumn{1}{c|}{0.074} &
  \multicolumn{1}{c|}{1.196} &
  \multicolumn{1}{c|}{0.869} &
  1.107 &
  91.73 (8, 576) \\ \cline{2-12} 
 &
  CT-AC$^\dagger$ &
  \multicolumn{1}{r|}{0.411} &
  \multicolumn{1}{r|}{1.000} &
  \multicolumn{1}{r|}{1.000} &
  \multicolumn{1}{r|}{-0.171} &
  -2061.6 &
  \multicolumn{1}{c|}{0.069} &
  \multicolumn{1}{c|}{\bf \blue{0.960}} &
  \multicolumn{1}{c|}{0.788} &
  {\bf \blue{0.943}} &
  89.83 (8, 576) \\ \cline{2-12} 
 
 &
  {\em SA} &
  \multicolumn{1}{r|}{0.408} &
  \multicolumn{1}{r|}{1.021} &
  \multicolumn{1}{r|}{0.994} &
  \multicolumn{1}{r|}{-0.186} &
  \blue{\bf -1499.3} &
  \multicolumn{1}{c|}{0.068} &
  \multicolumn{1}{c|}{1.020} &
  \multicolumn{1}{c|}{\bf \blue{0.756}} &
  0.956 &
  10.68 (0, 80) \\ \cline{2-12} 
 &
  RePlAce &
  \multicolumn{1}{r|}{0.409} &
  \multicolumn{1}{r|}{0.980} &
  \multicolumn{1}{r|}{0.982} &
  \multicolumn{1}{r|}{-0.209} &
  -1858.5 &
  \multicolumn{1}{c|}{0.059} &
  \multicolumn{1}{c|}{1.629} &
  \multicolumn{1}{c|}{1.250} &
  1.499 & 0.68 (0, 1)
   \\ \cline{2-12} 
 &
  CMP &
  \multicolumn{1}{r|}{0.405} &
  \multicolumn{1}{r|}{\blue{\bf 0.857}} &
  \multicolumn{1}{r|}{0.918} &
  \multicolumn{1}{r|}{-0.197} &
  -1961.3 &
  \multicolumn{1}{c|}{\bf \blue{0.059}} &
  \multicolumn{1}{c|}{1.526} &
  \multicolumn{1}{c|}{1.183} &
  1.413 & 1.19 (0, 8)
   \\ \cline{2-12} 
 &
  Human &
  \multicolumn{1}{r|}{0.406} &
  \multicolumn{1}{r|}{0.928} &
  \multicolumn{1}{r|}{0.944} &
  \multicolumn{1}{r|}{-0.149} &
  -1766.5 &
  \multicolumn{1}{c|}{0.069} &
  \multicolumn{1}{c|}{1.523} &
  \multicolumn{1}{c|}{1.278} &
  1.469 & NA
   \\ \hline
\end{tabular}%
% }
\end{table*}

\noindent
{\bf Evaluation of different macro placers on {\em CT-Ariane} and its scaled versions.}
Table~\ref{tab:ct_ariane} 
presents results for Google's public TSMC 7nm Ariane testcase ({\em CT-Ariane})
and its scaled (x2, x4) versions. Since the LEF/DEF files generated from
protobuf~\cite{CT-TCAD}, do not include the timing or layout information required for 
a complete P\&R flow, we only report post-detailed placement half-perimeter 
wirelength (DP-HPWL) along with proxy cost.\footnote{We run the {\em place\_design} command 
in Innovus 21.1 to place standard cells, then report ``Total half perimeter of net bounding box''
(DP-HPWL) from the log of the `earlyGlobalRoute' command.}
We observe the following.

\begin{itemize}[noitemsep, topsep=0pt, leftmargin=*]
    \item Both {\em CT-AC} and {\em CT-Scratch} produce better proxy cost (in all three components)
    than the reference (from Google's internal runs) provided in the {\em CT} repo \cite{CT-TCAD}.\footnote{We use 
    for comparison the best result (run\_07) reported in \cite{CT-TCAD}.} This
    indicates that we are training {\em CT-AC} and {\em CT-Scratch} correctly.
    
    \item {\em SA} dominates both {\em CT-AC} and {\em CT-Scratch} in terms of DP-HPWL and proxy cost,
     while using only a fraction of the runtime and compute resources.

    \item We follow the ``quantified scaling suboptimality'' methodology 
    in \cite{HagenHK95} to perform scaling studies of different macro placers.
    We define the {\em scaling suboptimality} $\alpha(k)$ as
    \begin{equation}
        \alpha(k) = \frac{\text{DP-HPWL}\_k}{(k \times \text{DP-HPWL}\_1)} - 1
    \end{equation}
    where $\text{DP-HPWL}\_k$ is the DP-HPWL of the scaled ($k$x) version of the base design.
    A lower value of $\alpha(k)$ indicates better scaling behavior. 
    For {\em CT-AC}, {\em SA}, RePlAce and Human solutions, ($\alpha(2)$, $\alpha(4)$)
    are  (0.000, 0.037), (0.009, 0.064), (0.005, -0.022) and (0.003, 0.028),
    respectively. 

\end{itemize}

\begin{table}[]
\caption{DP-HPWL and proxy cost, runtime, and resource details of
different macro placement solutions on {\em CT-Ariane} and its scaled (x2, x4) versions.
 $^\dagger$ indicates a total of 400
iterations used for training. Best DP-HPWL and proxy cost values are highlighted using blue bold font. \sk{\#G and \#C denote the number of V100 GPUs and CPU threads, respectively.}}
\label{tab:ct_ariane}
% \vspace{-0.05in}
{\centering
\resizebox{1.0\columnwidth}{!}{%
\begin{tabular}{|c|c|c|cccc|c|}
\hline
\multirow{2}{*}{Design} &
  \multirow{2}{*}{\begin{tabular}[c]{@{}c@{}}Macro \\ Placer\end{tabular}} &
  \multirow{2}{*}{\begin{tabular}[c]{@{}c@{}}DP-HPWL \\ ($\mu m$)\end{tabular}} &
  \multicolumn{4}{c|}{Proxy Cost Details} &
  \multirow{2}{*}{\begin{tabular}[c]{@{}c@{}}Runtime (Hrs) \\ (\#G, \#C)\end{tabular}} \\ \cline{4-7}
 &
   &
   &
  \multicolumn{1}{c|}{WL} &
  \multicolumn{1}{c|}{Den.} &
  \multicolumn{1}{c|}{Cong.} &
  Proxy &
   \\ \hline \hline
\multirow{7}{*}{\begin{tabular}[c]{@{}c@{}}CT-Ariane\end{tabular}} &
  CT repo~\cite{Ariane-pre-train} &
  NA &
  \multicolumn{1}{c|}{0.098} &
  \multicolumn{1}{c|}{0.511} &
  \multicolumn{1}{c|}{0.868} &
  0.787$^*$ &
  NA \\ \cline{2-8} 
 &
  CT-Scratch$^\dagger$ &
  958137 &
  \multicolumn{1}{c|}{0.094} &
  \multicolumn{1}{c|}{\bf \blue{0.503}} &
  \multicolumn{1}{c|}{0.854} &
  0.772 & 
  37.89 (8, 576) \\ \cline{2-8} 
 &
  CT-AC$^\dagger$ &
  938528 &
  \multicolumn{1}{c|}{0.092} &
  \multicolumn{1}{c|}{0.509} &
  \multicolumn{1}{c|}{0.829} &
  0.761 &
  55.16 (8, 576) \\ \cline{2-8} 
 &
  {\em SA} &
  804228 &
  \multicolumn{1}{c|}{\bf \blue{0.081}} &
  \multicolumn{1}{c|}{0.525} &
  \multicolumn{1}{c|}{\bf \blue{0.814}} &
  {\bf \blue{0.750}} &
  11.13 (0, 80) \\ \cline{2-8}
 &
  RePlAce &
  952429 &
  \multicolumn{1}{c|}{\bf \blue{0.081}} &
  \multicolumn{1}{c|}{0.992} &
  \multicolumn{1}{c|}{1.285} &
  1.219 & 0.03 (0, 1)
   \\ \cline{2-8} 
 &
  CMP &
  {\bf \blue{745370}} &
  \multicolumn{1}{c|}{0.075} &
  \multicolumn{1}{c|}{0.743} &
  \multicolumn{1}{c|}{0.999} &
  0.946 &
  0.02 (0, 16) \\ \cline{2-8} 
 &
  Human &
  931105 &
  \multicolumn{1}{c|}{0.093} &
  \multicolumn{1}{c|}{0.824} &
  \multicolumn{1}{c|}{1.241} &
  1.126 & NA
   \\ \hline \hline
\multirow{7}{*}{\begin{tabular}[c]{@{}c@{}}CT-Ariane\\-X2\end{tabular}} &
  CT-Scratch &
  2910809 &
  \multicolumn{1}{c|}{0.101} &
  \multicolumn{1}{c|}{0.533} &
  \multicolumn{1}{c|}{1.015} &
  0.876 &
  38.90 (8, 576) \\ \cline{2-8} 
 &
  CT-AC$^\dagger$ &
  1876365 &
  \multicolumn{1}{c|}{0.071} &
  \multicolumn{1}{c|}{0.493} &
  \multicolumn{1}{c|}{0.836} &
  0.735 &
  86.45 (8, 576) \\ \cline{2-8} 
 &
  {\em SA} &
  1623056 &
  \multicolumn{1}{c|}{\bf \blue{0.067}} &
  \multicolumn{1}{c|}{\bf \blue{0.490}} &
  \multicolumn{1}{c|}{\bf \blue{0.834}} &
  \blue{\bf 0.729} &
  6.93 (0, 80) \\ \cline{2-8} 
 &
  RePlAce &
  1913954 &
  \multicolumn{1}{c|}{0.078} &
  \multicolumn{1}{c|}{0.754} &
  \multicolumn{1}{c|}{1.091} &
  1.000 & 0.06 (0, 1)
   \\ \cline{2-8} 
 &
  CMP &
  \blue{\bf 1510219} &
  \multicolumn{1}{c|}{0.074} &
  \multicolumn{1}{c|}{0.620} &
  \multicolumn{1}{c|}{1.134} &
  0.951 &
  0.04 (0, 16) \\ \cline{2-8} 
  &
  Human &
  1868380 &
  \multicolumn{1}{c|}{0.081} &
  \multicolumn{1}{c|}{0.832} &
  \multicolumn{1}{c|}{1.227} &
  1.111 & NA
   \\ \hline \hline
\multirow{7}{*}{\begin{tabular}[c]{@{}c@{}}CT-Ariane\\-X4\end{tabular}} &
  CT-Scratch &
  \multicolumn{6}{c|}{Divergence} \\ \cline{2-8} 
 &
  CT-AC$^\dagger$ &
  3893091 &
  \multicolumn{1}{c|}{0.056} &
  \multicolumn{1}{c|}{\blue{\bf 0.466}} &
  \multicolumn{1}{c|}{0.836} &
  0.707 &
  188.69 (8, 576) \\ \cline{2-8} 
 &
  CT-Ours$^\dagger$ &
  5525222 &
  \multicolumn{1}{c|}{0.075} &
  \multicolumn{1}{c|}{0.473} &
  \multicolumn{1}{c|}{0.922} &
   0.772&
  174.07 (8, 576) \\ \cline{2-8}   
 
 &
  {\em SA} &
  3423907 &
  \multicolumn{1}{c|}{\blue{\bf 0.052}} &
  \multicolumn{1}{c|}{0.467} &
  \multicolumn{1}{c|}{\blue{\bf 0.815}} &
  \blue{\bf 0.693} &
  9.97 (0, 80) \\ \cline{2-8} 
 &
  RePlAce &
  3726672 &
  \multicolumn{1}{c|}{0.055} &
  \multicolumn{1}{c|}{0.730} &
  \multicolumn{1}{c|}{1.062} &
  0.950 &
  0.17 (0, 1) \\ \cline{2-8} 
 &
  CMP &
  \blue{\bf 3049402} &
  \multicolumn{1}{c|}{0.055} &
  \multicolumn{1}{c|}{0.735} &
  \multicolumn{1}{c|}{1.139} &
  0.992 &
  0.08 (0, 16) \\ \cline{2-8} 
 &
  Human &
  3827163 &
  \multicolumn{1}{c|}{0.060} &
  \multicolumn{1}{c|}{0.757} &
  \multicolumn{1}{c|}{1.590} &
  1.233 & NA
   \\ \hline
\end{tabular}%
}
}
% \scriptsize
\vspace{0.01in} \\
%\small
\footnotesize
$^*$We compute the proxy cost for {\em CT-Ariane} using the cost components
reported in the {\em CT} repo~\cite{Ariane-pre-train}, with weights of 1.0 for wirelength
and 0.5 for both density and congestion.

\end{table}

\subsection{Stability studies}
\label{subsec:stability}
Outcomes from {\em CT} training, {\em SA} execution, and hMETIS-based {\em grouping}
will all vary according to given input seeds.\footnote{\label{fn:seed}Perturbing
the design (e.g., changing the SDC clock period by $\pm 1$
ps or the floorplan width by $\pm 1$ site) produces different outputs
from CMP and RePlAce, thereby modifying the input design.
We evaluate seed-induced variation in {\em SA} and {\em CT} while holding
the input design fixed. Because CMP and RePlAce do not expose a seed and
produce deterministic outputs for a fixed input design, we do not perform
variation studies for them.
}
We have studied how this can
affect final proxy cost and postRouteOpt PPA metrics. In the following,
we present stability issues observed in {\em CT}, then analyze seed effects
in {\em CT-Scratch}. We then examine seed effects on {\em SA},
and seed effects on hMETIS-based grouping.

\begin{figure}[ht]
    \centering
    \includegraphics[width=0.49\columnwidth]{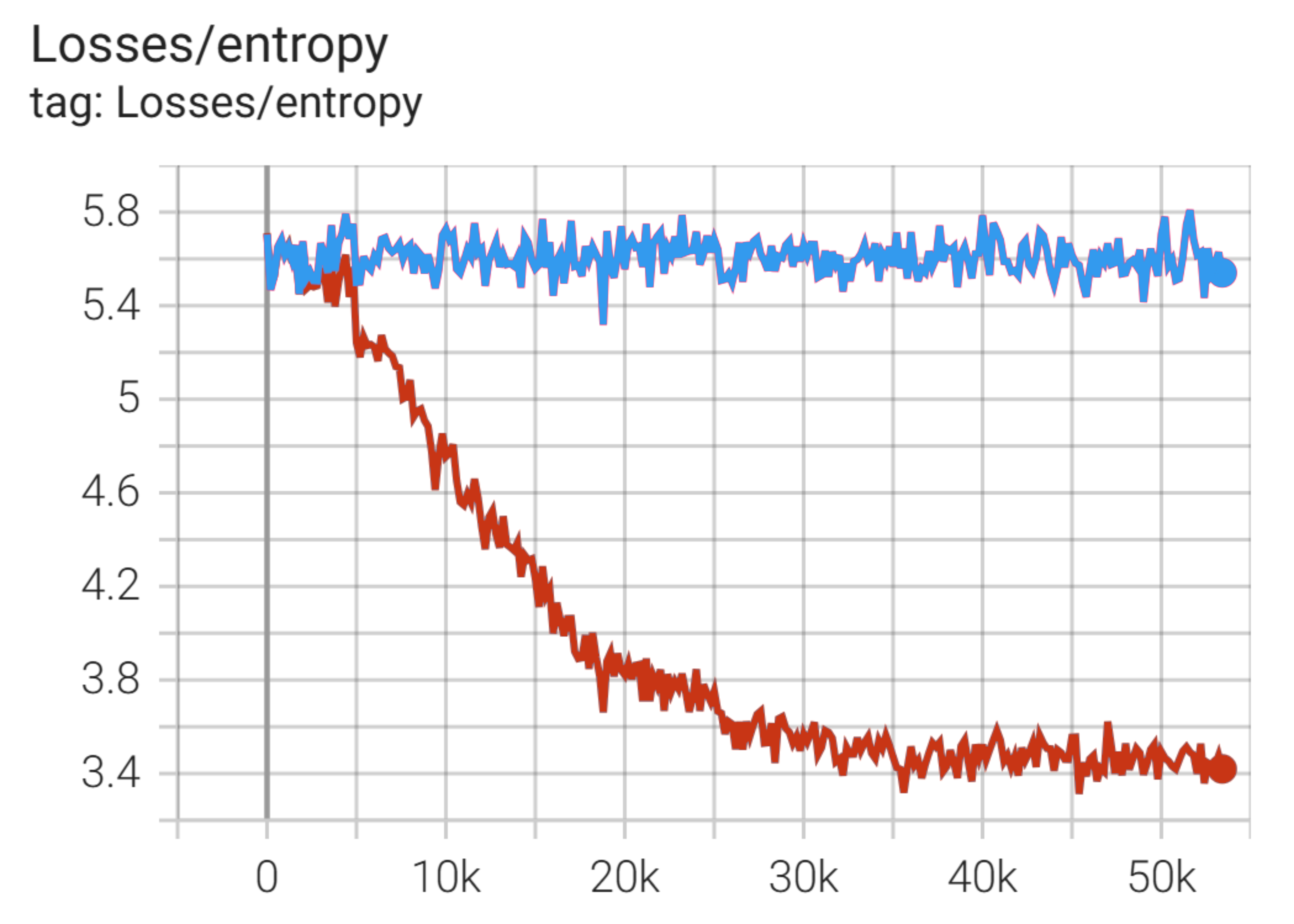}
    \includegraphics[width=0.49\columnwidth]{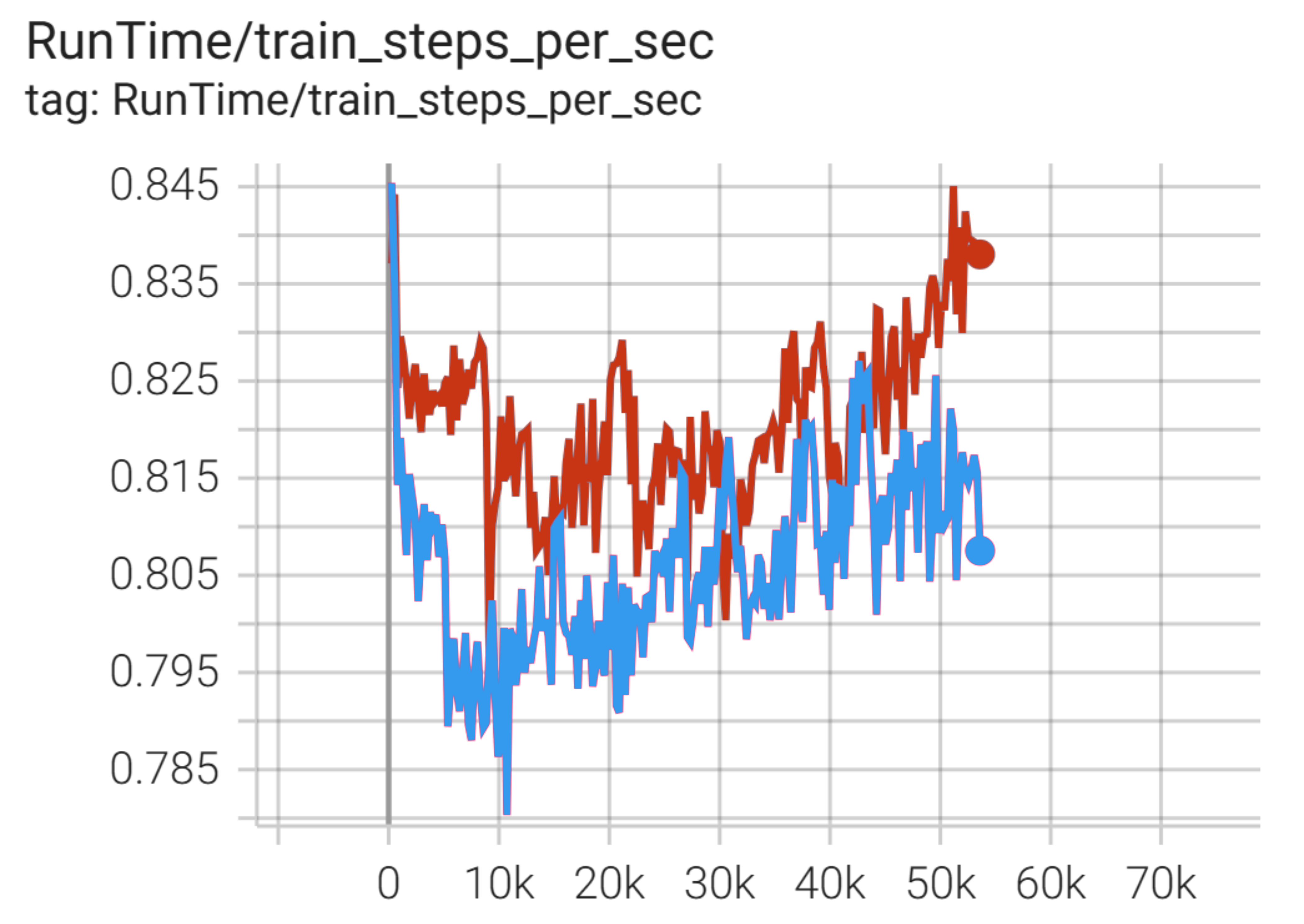}
    \caption{Loss (left) and train steps per second (right) for converged 
    (red) and diverged (blue) {\em CT} runs on Ariane-GF12. 
    Both runs use the same machine, same environment, same netlist,
    and same (default) seed = 333.}
    \label{fig:gf12-instability}
\end{figure}

\noindent
{\bf Stability issues of {\em CT}.}
We observe that on the same machine, in the same environment, and for the same
netlist and seed (e.g., the default ``333''), {\em CT} can converge in one run 
but diverge in another. 
Further, all of our {\em CT} runs use machines with identical configurations, 
and we also observe different outcomes from otherwise identical runs
that are executed on different machines, regardless of netlist size.
Figure~\ref{fig:gf12-instability} plots loss and train steps
per second for the Ariane-GF12 design, where two identically-configured
runs result in convergence (red) and divergence (blue). 
Although both runs reach a similar speed of around 0.8 steps per second, the loss 
does not decrease for the diverged run.
Given this behavior, if a {\em CT} run diverges, more runs must be launched 
to confirm divergence. Determining exact causes and resource-efficient
mitigations of this stochasticity is an open direction for future investigation.

\noindent
{\bf Seed effects in {\em CT}.} 
Despite non-determinism of {\em CT} training and its outcomes (even with
the same seed, environment, and machine), we have sought to understand
the variability of {\em CT} outcomes, as follows.
(i) We set three global seeds (111, 222, and 333) for random weight 
initialization. (ii) We then perform three independent runs
of {\em CT-Scratch} for 400 iterations on the Ariane-ASAP7 design {\em per
each seed}, resulting in a total of nine runs. 
Table~\ref{tab:ct_stability} shows final proxy cost and postRouteOpt PPA of 
these nine runs.\footnote{We note that {\em Nature} Extended Data Table 5
~\cite{MirhoseiniGYJSWLJPNPTSHTLLHCD21} and \cite{PeerReviewNature} discuss
seed effects on variation, but not the stochasticity effects that we document here.}

\begin{table}[]
% \caption{Mean (standard deviation) of proxy cost and {\em Nature} {\em Table 1} 
% metrics for different global seeds in {\em CT}.}
\caption{Variation in postRouteOpt PPA and proxy cost for {\em CT-Scratch} on 
Ariane-ASAP7 across 9 runs (3 global seeds $\times$ 3 runs per
seed).}
\label{tab:ct_stability}
% \vspace{-0.05in}
\resizebox{\columnwidth}{!}{%
\begin{tabular}{|c|ccccc|cccc|}
\hline
\multirow{2}{*}{Seed} &
  \multicolumn{5}{c|}{PostRouteOpt PPA  (From Innovus)} &
  \multicolumn{4}{c|}{Proxy Cost Details} \\ \cline{2-10} 
 &
  \multicolumn{1}{c|}{\begin{tabular}[c]{@{}c@{}}Area \\ ($\mu m^2$)\end{tabular}} &
  \multicolumn{1}{c|}{\begin{tabular}[c]{@{}c@{}}rWL \\ ($\mu m$)\end{tabular}} &
  \multicolumn{1}{c|}{\begin{tabular}[c]{@{}c@{}}Power \\ ($mW$)\end{tabular}} &
  \multicolumn{1}{c|}{\begin{tabular}[c]{@{}c@{}}WNS \\ ($ps$)\end{tabular}} &
  \begin{tabular}[c]{@{}c@{}}TNS \\ ($ns$)\end{tabular} &
  \multicolumn{1}{c|}{WL} &
  \multicolumn{1}{c|}{Den.} &
  \multicolumn{1}{c|}{Cong.} &
  Proxy \\ \hline \hline
\multirow{3}{*}{111} &
  \multicolumn{1}{c|}{16528} &
  \multicolumn{1}{c|}{1022767} &
  \multicolumn{1}{c|}{505.6} &
  \multicolumn{1}{c|}{-128} &
  -133.6 &
  \multicolumn{1}{c|}{0.123} &
  \multicolumn{1}{c|}{0.802} &
  \multicolumn{1}{c|}{0.848} &
  0.947 \\ \cline{2-10}

 &
  \multicolumn{1}{c|}{16419} &
  \multicolumn{1}{c|}{991490} &
  \multicolumn{1}{c|}{504.1} &
  \multicolumn{1}{c|}{-97} &
  -59.9 &
  \multicolumn{1}{c|}{0.118} &
  \multicolumn{1}{c|}{0.809} &
  \multicolumn{1}{c|}{0.822} &
  0.934 \\ \cline{2-10} 
  
 &
  \multicolumn{1}{c|}{16410} &
  \multicolumn{1}{c|}{993215} &
  \multicolumn{1}{c|}{504.6} &
  \multicolumn{1}{c|}{-122} &
  -123.8 &
  \multicolumn{1}{c|}{0.121} &
  \multicolumn{1}{c|}{0.809} &
  \multicolumn{1}{c|}{0.842} &
  0.946 \\ \hline \hline

\multirow{3}{*}{222} &
  \multicolumn{1}{c|}{16446} &
  \multicolumn{1}{c|}{1014526} &
  \multicolumn{1}{c|}{502.0} &
  \multicolumn{1}{c|}{-112} &
  -84.6 &
  \multicolumn{1}{c|}{0.120} &
  \multicolumn{1}{c|}{0.807} &
  \multicolumn{1}{c|}{0.834} &
  0.940 \\ \cline{2-10} 
 &
  \multicolumn{1}{c|}{16537} &
  \multicolumn{1}{c|}{985154} &
  \multicolumn{1}{c|}{505.7} &
  \multicolumn{1}{c|}{-142} &
  -163.4 &
  \multicolumn{1}{c|}{0.119} &
  \multicolumn{1}{c|}{0.825} &
  \multicolumn{1}{c|}{0.828} &
  0.945 \\ \cline{2-10} 
 &
  \multicolumn{1}{c|}{16482} &
  \multicolumn{1}{c|}{1006802} &
  \multicolumn{1}{c|}{505.3} &
  \multicolumn{1}{c|}{-131} &
  -127.4 &
  \multicolumn{1}{c|}{0.119} &
  \multicolumn{1}{c|}{0.805} &
  \multicolumn{1}{c|}{0.843} &
  0.943 \\ \hline \hline
\multirow{3}{*}{333} &
  \multicolumn{1}{c|}{16528} &
  \multicolumn{1}{c|}{1013108} &
  \multicolumn{1}{c|}{504.5} &
  \multicolumn{1}{c|}{-128} &
  -164.1 &
  \multicolumn{1}{c|}{0.119} &
  \multicolumn{1}{c|}{0.821} &
  \multicolumn{1}{c|}{0.871} &
  0.965 \\ \cline{2-10} 
 &
  \multicolumn{1}{c|}{16630} &
  \multicolumn{1}{c|}{1016008} &
  \multicolumn{1}{c|}{506.4} &
  \multicolumn{1}{c|}{-162} &
  -228.0 &
  \multicolumn{1}{c|}{0.126} &
  \multicolumn{1}{c|}{0.801} &
  \multicolumn{1}{c|}{0.855} &
  0.953 \\ \cline{2-10} 
 &
  \multicolumn{1}{c|}{16602} &
  \multicolumn{1}{c|}{1026772} &
  \multicolumn{1}{c|}{506.1} &
  \multicolumn{1}{c|}{-198} &
  -355.0 &
  \multicolumn{1}{c|}{0.125} &
  \multicolumn{1}{c|}{0.829} &
  \multicolumn{1}{c|}{0.851} &
  0.964 \\ \hline \hline
Mean &
  \multicolumn{1}{c|}{16509} &
  \multicolumn{1}{c|}{1007760} &
  \multicolumn{1}{c|}{504.9} &
  \multicolumn{1}{c|}{-136} &
  -160.0 &
  \multicolumn{1}{c|}{0.121} &
  \multicolumn{1}{c|}{0.812} &
  \multicolumn{1}{c|}{0.844} &
  0.949 \\ \hline
STD &
  \multicolumn{1}{c|}{76.8} &
  \multicolumn{1}{c|}{14651.3} &
  \multicolumn{1}{c|}{1.34} &
  \multicolumn{1}{c|}{29.5} &
  87.65 &
  \multicolumn{1}{c|}{0.003} &
  \multicolumn{1}{c|}{0.010} &
  \multicolumn{1}{c|}{0.015} &
  0.010 \\ \hline
\end{tabular}%
}
\end{table}

\noindent
{\bf Seed effects in {\em SA}.}
{\em SA} runs are deterministic: when run with the same seed,
on the same machine and in the same environment, {\em SA}
produces the exact same result. 
Table~\ref{tab:sa_seed} gives details of proxy cost
and postRouteOpt metrics for the Ariane-ASAP7 design when {\em SA}
is run with ten different seeds.
From Tables~\ref{tab:ct_stability} and ~\ref{tab:sa_seed},
we observe the following.
\begin{itemize}[noitemsep, topsep=0pt, leftmargin=*]
    \item For proxy cost, {\em CT-Scratch} and {\em SA} show 
    very similar variation, which differs from the observation in~\cite{ispd23}. 
    Decreased variation of proxy cost for {\em CT-Scratch} may be due 
    in part to updates in the new version of {\em CT} and to our training 
    for 400 iterations, as opposed to 200 in  \cite{ispd23}. 
    Increased variation in {\em SA} proxy cost may be due to use
    of 80 {\em SA} workers, as opposed to 320 workers in~\cite{ispd23}.
    \item For PPA (rWL, power and TNS), {\em SA} shows significantly
    less variation than {\em CT-Scratch}. Furthermore, mean PPA and 
    proxy cost values from {\em SA} are better 
    than those from {\em CT-Scratch}, indicating that {\em SA} outperforms {\em CT-Scratch}.
% Alternate: Mean PPA and proxy cost values show statistically significant differences, indicating that {\em SA} outperforms {\em CT-Scratch}.
\end{itemize}

\begin{table}[]
\caption{Variation in postRouteOpt PPA and proxy cost for {\em SA} on the
Ariane-ASAP7 design across 10 seeds.}
\label{tab:sa_seed}
% \vspace{-0.05in}
\resizebox{\columnwidth}{!}{%
\begin{tabular}{|c|ccccc|cccc|}
\hline
\multirow{2}{*}{Seed} &
  \multicolumn{5}{c|}{PostRouteOpt PPA  (From Innovus)} &
  \multicolumn{4}{c|}{Proxy Cost Details} \\ \cline{2-10} 
 &
  \multicolumn{1}{c|}{\begin{tabular}[c]{@{}c@{}}Area \\ ($\mu m^2$)\end{tabular}} &
  \multicolumn{1}{c|}{\begin{tabular}[c]{@{}c@{}}rWL \\ ($\mu m$)\end{tabular}} &
  \multicolumn{1}{c|}{\begin{tabular}[c]{@{}c@{}}Power \\ ($mW$)\end{tabular}} &
  \multicolumn{1}{c|}{\begin{tabular}[c]{@{}c@{}}WNS \\ ($ps$)\end{tabular}} &
  \begin{tabular}[c]{@{}c@{}}TNS \\ ($ns$)\end{tabular} &
  \multicolumn{1}{c|}{WL} &
  \multicolumn{1}{c|}{Den.} &
  \multicolumn{1}{c|}{Cong.} &
  Proxy \\ \hline \hline
1 &
  \multicolumn{1}{c|}{16467} &
  \multicolumn{1}{c|}{886776} &
  \multicolumn{1}{c|}{503.5} &
  \multicolumn{1}{c|}{-124} &
  -141.1 &
  \multicolumn{1}{c|}{0.108} &
  \multicolumn{1}{c|}{0.817} &
  \multicolumn{1}{c|}{0.822} &
  0.928 \\ \hline
111 &
  \multicolumn{1}{c|}{16503} &
  \multicolumn{1}{c|}{902468} &
  \multicolumn{1}{c|}{503.6} &
  \multicolumn{1}{c|}{-120} &
  -131.3 &
  \multicolumn{1}{c|}{0.111} &
  \multicolumn{1}{c|}{0.820} &
  \multicolumn{1}{c|}{0.845} &
  0.943 \\ \hline
222 &
  \multicolumn{1}{c|}{16380} &
  \multicolumn{1}{c|}{901541} &
  \multicolumn{1}{c|}{500.7} &
  \multicolumn{1}{c|}{-89} &
  -62.1 &
  \multicolumn{1}{c|}{0.108} &
  \multicolumn{1}{c|}{0.817} &
  \multicolumn{1}{c|}{0.802} &
  0.917 \\ \hline
333 &
  \multicolumn{1}{c|}{16494} &
  \multicolumn{1}{c|}{896866} &
  \multicolumn{1}{c|}{504.7} &
  \multicolumn{1}{c|}{-102} &
  -83.7 &
  \multicolumn{1}{c|}{0.108} &
  \multicolumn{1}{c|}{0.803} &
  \multicolumn{1}{c|}{0.848} &
  0.934 \\ \hline
444 &
  \multicolumn{1}{c|}{16401} &
  \multicolumn{1}{c|}{900137} &
  \multicolumn{1}{c|}{503.9} &
  \multicolumn{1}{c|}{-70} &
  -37.3 &
  \multicolumn{1}{c|}{0.108} &
  \multicolumn{1}{c|}{0.818} &
  \multicolumn{1}{c|}{0.812} &
  0.923 \\ \hline
555 &
  \multicolumn{1}{c|}{16461} &
  \multicolumn{1}{c|}{894386} &
  \multicolumn{1}{c|}{504.1} &
  \multicolumn{1}{c|}{-81} &
  -56.9 &
  \multicolumn{1}{c|}{0.105} &
  \multicolumn{1}{c|}{0.814} &
  \multicolumn{1}{c|}{0.813} &
  0.919 \\ \hline
666 &
  \multicolumn{1}{c|}{16420} &
  \multicolumn{1}{c|}{898643} &
  \multicolumn{1}{c|}{504.3} &
  \multicolumn{1}{c|}{-109} &
  -95.7 &
  \multicolumn{1}{c|}{0.107} &
  \multicolumn{1}{c|}{0.802} &
  \multicolumn{1}{c|}{0.808} &
  0.912 \\ \hline
777 &
  \multicolumn{1}{c|}{16563} &
  \multicolumn{1}{c|}{898613} &
  \multicolumn{1}{c|}{504.3} &
  \multicolumn{1}{c|}{-118} &
  -113.1 &
  \multicolumn{1}{c|}{0.107} &
  \multicolumn{1}{c|}{0.815} &
  \multicolumn{1}{c|}{0.835} &
  0.933 \\ \hline
888 &
  \multicolumn{1}{c|}{16486} &
  \multicolumn{1}{c|}{898399} &
  \multicolumn{1}{c|}{504.8} &
  \multicolumn{1}{c|}{-133} &
  -110.3 &
  \multicolumn{1}{c|}{0.106} &
  \multicolumn{1}{c|}{0.822} &
  \multicolumn{1}{c|}{0.806} &
  0.920 \\ \hline
999 &
  \multicolumn{1}{c|}{16522} &
  \multicolumn{1}{c|}{904774} &
  \multicolumn{1}{c|}{502.6} &
  \multicolumn{1}{c|}{-121} &
  -105.3 &
  \multicolumn{1}{c|}{0.107} &
  \multicolumn{1}{c|}{0.813} &
  \multicolumn{1}{c|}{0.810} &
  0.918 \\ \hline \hline
Mean &
  \multicolumn{1}{c|}{16470} &
  \multicolumn{1}{c|}{898260} &
  \multicolumn{1}{c|}{503.7} &
  \multicolumn{1}{c|}{-105} &
  -93.7 &
  \multicolumn{1}{c|}{0.107} &
  \multicolumn{1}{c|}{0.815} &
  \multicolumn{1}{c|}{0.822} &
  0.926 \\ \hline
STD &
  \multicolumn{1}{c|}{56.4} &
  \multicolumn{1}{c|}{4984.1} &
  \multicolumn{1}{c|}{1.21} &
  \multicolumn{1}{c|}{20.6} &
  33.47 &
  \multicolumn{1}{c|}{0.002} &
  \multicolumn{1}{c|}{0.007} &
  \multicolumn{1}{c|}{0.017} &
  0.010 \\ \hline
\end{tabular}%
}
\end{table}

\noindent
{\bf Seed effects in CT-Grouping.}
The CT-Grouping flow in {\em CT} uses the {\em hMETIS} binary, which 
does not support a seed input and is non-deterministic. An example
consequence is that we cannot reproduce the generation of
the clustered netlists 
used in~\cite{ispd23}. However, the {\em hMETIS} shared library
C API does support specification of a seed: we have written 
a C++ wrapper (publicly available in~\cite{TILOS-repo}) which 
accepts a seed and  ensures reproducibility of the grouping flow. 
To study the effect of seeds in grouping, we generate five
clustered netlists for Ariane-ASAP7 using seeds 
(${111, 222, \dots, 555}$); these netlists respectively
have 782, 791, 784, 785 and 786 standard-cell clusters.
We then run {\em SA} and evaluate the macro placement solutions
using the evaluation flow described in Subsection~\ref{subsec:evalflow}. 
Table~\ref{tab:grp_seed} presents the postRouteOpt PPA and proxy cost 
for these five runs.  We see that variation in proxy cost is 
similar to that seen in the seed study of {\em SA} 
(Table~\ref{tab:sa_seed}), while there is larger variation in 
postRouteOpt PPA metrics.

Further, we study the {\em combined} effects from seeding of CT-Grouping 
and seeding of {\em SA}, by ``crossing''
the first six rows of Table~\ref{tab:sa_seed} with 
the five rows of Table~\ref{tab:grp_seed}.
That is, for each of the five clustered netlists 
generated using different grouping seeds, we run 
six {\em SA} runs with different seeds, and capture 
resulting postRouteOpt PPA and proxy costs.
Mean (\sk{standard deviation, STD}) for power, WNS, TNS and proxy cost are respectively 504.3 (0.69),
-117 (59.1), -106.5 (52.27) and 0.929 (0.009). For proxy cost, the combined effect of
{\em SA} and grouping is similar to the effect of only grouping or only {\em SA}. 
For PPA, the combined effect shows less variation in power,
WNS and TNS compared to varying only the grouping seed. 
Compared to varying only the {\em SA} seed, variation in TNS and WNS increases while 
variation in power decreases.

\begin{table}[]
\caption{Variation in postRouteOpt PPA and proxy cost for {\em SA} on 5 
different clustered netlists of the Ariane-ASAP7 design, when hMETIS is
run with 5 different seeds in the {\em CT} grouping flow.}
\label{tab:grp_seed}
% \vspace{-0.05in}
\resizebox{\columnwidth}{!}{%
\begin{tabular}{|c|ccccc|cccc|}
\hline
\multirow{3}{*}{\begin{tabular}[c]{@{}c@{}}GRP \\ Seed\end{tabular}} &
  \multicolumn{5}{c|}{PostRouteOpt PPA  (From Innovus)} &
  \multicolumn{4}{c|}{Proxy Cost Details} \\ \cline{2-10} 
 &
  \multicolumn{1}{c|}{\begin{tabular}[c]{@{}c@{}}Area \\ ($\mu m^2$)\end{tabular}} &
  \multicolumn{1}{c|}{\begin{tabular}[c]{@{}c@{}}rWL \\ ($\mu m$)\end{tabular}} &
  \multicolumn{1}{c|}{\begin{tabular}[c]{@{}c@{}}Power \\ ($mW$)\end{tabular}} &
  \multicolumn{1}{c|}{\begin{tabular}[c]{@{}c@{}}WNS \\ ($ps$)\end{tabular}} &
  \begin{tabular}[c]{@{}c@{}}TNS \\ ($ns$)\end{tabular} &
  \multicolumn{1}{c|}{WL} &
  \multicolumn{1}{c|}{Den.} &
  \multicolumn{1}{c|}{Cong.} &
  Proxy \\ \hline \hline
111 &
  \multicolumn{1}{c|}{16412} &
  \multicolumn{1}{c|}{896408} &
  \multicolumn{1}{c|}{502.6} &
  \multicolumn{1}{c|}{-97} &
  -74.8 &
  \multicolumn{1}{c|}{0.104} &
  \multicolumn{1}{c|}{0.830} &
  \multicolumn{1}{c|}{0.806} &
  0.922 \\ \hline
222 &
  \multicolumn{1}{c|}{16434} &
  \multicolumn{1}{c|}{886847} &
  \multicolumn{1}{c|}{504.3} &
  \multicolumn{1}{c|}{-104} &
  -72.7 &
  \multicolumn{1}{c|}{0.107} &
  \multicolumn{1}{c|}{0.815} &
  \multicolumn{1}{c|}{0.817} &
  0.923 \\ \hline
333 &
  \multicolumn{1}{c|}{16448} &
  \multicolumn{1}{c|}{907544} &
  \multicolumn{1}{c|}{503.6} &
  \multicolumn{1}{c|}{-95} &
  -91.8 &
  \multicolumn{1}{c|}{0.104} &
  \multicolumn{1}{c|}{0.824} &
  \multicolumn{1}{c|}{0.802} &
  0.917 \\ \hline
444 &
  \multicolumn{1}{c|}{16431} &
  \multicolumn{1}{c|}{915978} &
  \multicolumn{1}{c|}{504.4} &
  \multicolumn{1}{c|}{-104} &
  -98.5 &
  \multicolumn{1}{c|}{0.104} &
  \multicolumn{1}{c|}{0.843} &
  \multicolumn{1}{c|}{0.790} &
  0.920 \\ \hline
555 &
  \multicolumn{1}{c|}{16433} &
  \multicolumn{1}{c|}{904477} &
  \multicolumn{1}{c|}{503.9} &
  \multicolumn{1}{c|}{-398} &
  -219.7 &
  \multicolumn{1}{c|}{0.109} &
  \multicolumn{1}{c|}{0.850} &
  \multicolumn{1}{c|}{0.806} &
  0.937 \\ \hline \hline
Mean &
  \multicolumn{1}{c|}{16432} &
  \multicolumn{1}{c|}{902251} &
  \multicolumn{1}{c|}{503.7} &
  \multicolumn{1}{c|}{-160} &
  -111.5 &
  \multicolumn{1}{c|}{0.106} &
  \multicolumn{1}{c|}{0.832} &
  \multicolumn{1}{c|}{0.804} &
  0.924 \\ \hline
STD &
  \multicolumn{1}{c|}{12.8} &
  \multicolumn{1}{c|}{11100.2} &
  \multicolumn{1}{c|}{0.70} &
  \multicolumn{1}{c|}{133.3} &
  61.47 &
  \multicolumn{1}{c|}{0.002} &
  \multicolumn{1}{c|}{0.014} &
  \multicolumn{1}{c|}{0.010} &
  0.008 \\ \hline
\end{tabular}%
}
\end{table}

\subsection{Ablation studies}
\label{subsec:ablation}
We present results and takeaways from  
ablation, stability and related studies; 
further examples appear in
\cite{TILOS-repo} \cite{OurProgress}.

\noindent
{\bf Effect of initial placement on the {\em CT} outcome.}
{\em CT}'s use of a physical synthesis tool (Synopsys DC-Topographical)
and the initial placement locations that it outputs with the gate-level
netlist is well-discussed in~\cite{PeerReviewNature} \cite{GoldieMYJSWLJPNPTSHTLLHCD24}. 
To evaluate the effect of initial placement, we generate four new
clustered netlists for the Ariane-ASAP7 testcase when all cells are
placed (i) at the lower left corner (0, 0), (ii) in the middle of the
canvas (111.204, 110.970), (iii) at the upper right corner (222.408,
221.940), and (iv) with no placement information.\footnote{To model
the absence of placement information, we turn off the {\em breakup} 
flag in the CT grouping flow.} We run CT from scratch for 400 iterations 
to obtain a macro placement for each clustered netlist,
then run our evaluation flow to capture postRouteOpt PPA. 
Table~\ref{tab:init_placement} shows that PPA numbers are very similar
to those from the default run -- which uses placement information from
Genus iSpatial. This observation differs from that of~\cite{ispd23}, 
which found up to 10\% improvement in rWL when a Genus iSpatial 
placement solution was used. (\cite{ispd23} used the older 
version of CT, trained for 200 iterations, and tested on the 
Ariane-NG45 design.)

\begin{table}[]
\caption{Effect of initial placement on {\em CT-Scratch} outcome for Ariane-ASAP7.}
\label{tab:init_placement}
\vspace{-0.05in}
\resizebox{\columnwidth}{!}{%
\begin{tabular}{|c|ccccc|cccc|}
\hline
\multirow{2}{*}{\begin{tabular}[c]{@{}c@{}}Initial\\ Placement\end{tabular}} &
  \multicolumn{5}{c|}{PostRouteOpt PPA  (From Innovus)} &
  \multicolumn{4}{c|}{Proxy Cost Details} \\ \cline{2-10} 
 &
  \multicolumn{1}{c|}{\begin{tabular}[c]{@{}c@{}}Area \\ ($\mu m^2$)\end{tabular}} &
  \multicolumn{1}{c|}{\begin{tabular}[c]{@{}c@{}}rWL \\ ($\mu m$)\end{tabular}} &
  \multicolumn{1}{c|}{\begin{tabular}[c]{@{}c@{}}Power \\ ($mW$)\end{tabular}} &
  \multicolumn{1}{c|}{\begin{tabular}[c]{@{}c@{}}WNS \\ ($ps$)\end{tabular}} &
  \begin{tabular}[c]{@{}c@{}}TNS \\ ($ns$)\end{tabular} &
  \multicolumn{1}{c|}{WL} &
  \multicolumn{1}{c|}{Den.} &
  \multicolumn{1}{c|}{Cong.} &
  Proxy \\ \hline \hline
Genus iSpatial &
  \multicolumn{1}{c|}{16570} &
  \multicolumn{1}{c|}{1026239} &
  \multicolumn{1}{c|}{505.4} &
  \multicolumn{1}{c|}{-142} &
  -184.2 &
  \multicolumn{1}{c|}{0.119} &
  \multicolumn{1}{c|}{0.821} &
  \multicolumn{1}{c|}{0.871} &
  0.965 \\ \hline
Center &
  \multicolumn{1}{c|}{16444} &
  \multicolumn{1}{c|}{984576} &
  \multicolumn{1}{c|}{504.3} &
  \multicolumn{1}{c|}{-108} &
  -79.0 &
  \multicolumn{1}{c|}{0.119} &
  \multicolumn{1}{c|}{0.789} &
  \multicolumn{1}{c|}{0.840} &
  0.933 \\ \hline
Lower Left &
  \multicolumn{1}{c|}{16466} &
  \multicolumn{1}{c|}{951872} &
  \multicolumn{1}{c|}{503.6} &
  \multicolumn{1}{c|}{-121} &
  -113.2 &
  \multicolumn{1}{c|}{0.110} &
  \multicolumn{1}{c|}{0.795} &
  \multicolumn{1}{c|}{0.834} &
  0.925 \\ \hline
Upper Right &
  \multicolumn{1}{c|}{16416} &
  \multicolumn{1}{c|}{992912} &
  \multicolumn{1}{c|}{504.7} &
  \multicolumn{1}{c|}{-105} &
  -83.0 &
  \multicolumn{1}{c|}{0.115} &
  \multicolumn{1}{c|}{0.821} &
  \multicolumn{1}{c|}{0.829} &
  0.939 \\ \hline
No Placement &
  \multicolumn{1}{c|}{16399} &
  \multicolumn{1}{c|}{975230} &
  \multicolumn{1}{c|}{502.3} &
  \multicolumn{1}{c|}{-98} &
  -81.9 &
  \multicolumn{1}{c|}{0.129} &
  \multicolumn{1}{c|}{0.843} &
  \multicolumn{1}{c|}{0.870} &
  0.986 \\ \hline
\end{tabular}%
}
\end{table}

\noindent
{\bf Correlation of proxy cost to {\em Nature Table 1} metrics.}
The RL agent in {\em Nature} and {\em CT} is driven by proxy cost,
while the EDA tool's post-P\&R output is ``ground truth''~\cite{MirhoseiniGYJSWLJPNPTSHTLLHCD21}. 
To study correlation of proxy cost with {\em Nature} {\em Table 1}
metrics, we collect 30 macro placements produced by 
{\em CT-Scratch} for Ariane-ASAP7 having proxy cost less than 1.0, 
and generate {\em Table 1} metrics for each.
Table~\ref{tab:correlation} shows the Kendall rank correlation
of proxy cost and {\em Table 1} metrics. Values close 
to +1 (resp. -1) indicate strong correlation (resp. anticorrelation), 
while values close to 0 indicate lack of correlation.
Similar to \cite{ispd23}, in the regime 
of relatively low proxy cost (high solution  quality), we observe 
poor correlation of proxy cost and its components with {\em Table 1} metrics. 
Compared to the proxy cost correlation study on Ariane-NG45 
in~\cite{ispd23} (Table 2),
we see that the new {\em CT-Scratch} on Ariane-ASAP7 shows 
improved correlation between proxy cost and rWL, but degraded 
correlation between proxy cost and WNS. Overall, the proxy cost
function optimized in the {\em Nature} paper shows poor correlation
with final chip metrics; therefore, relying on this proxy as an
optimization target is not a sound physical design strategy.

\begin{table}[]
\caption{Kendall rank correlation coefficient between proxy cost and 
{\em Nature Table 1} metrics for Ariane-ASAP7.}
\vspace{-0.05in}
\label{tab:correlation}
\centering
\resizebox{0.65\columnwidth}{!} {
\begin{tabular}{|c|c|c|c|c|c|}
\hline
\makecell{ Proxy Cost \\  Element} & \multicolumn{1}{c|}{Area} 
    & \multicolumn{1}{c|}{rWL} 
    & \multicolumn{1}{c|}{ Power} 
    & \multicolumn{1}{c|}{WNS} & \multicolumn{1}{c|}{TNS} \\ \hline
Wirelength    & 0.046
            & 0.355
            & 0.222 
            & -0.002 
            & 0.019 \\ \hline
Congestion    
        & 0.053 
        & 0.297 
        & 0.251 
        & 0.116 
        & 0.149
        \\ \hline
Density       
        & -0.055 
        & 0.299 
        & 0.299 
        & -0.039 
        & 0.023     \\ \hline
Proxy         
        & 0.058 
        & 0.402 
        & 0.320 
        & 0.051 
        & 0.090  \\ \hline
\end{tabular}
}
\end{table}

\noindent
{\bf Effect of different Innovus versions.}
We have studied whether differences between CMP versions affect
our conclusions. We run CMP in Cadence Innovus versions 
19.1, 20.1, and 21.1 to obtain three macro placement solutions
for the Ariane-ASAP7 design. We then run the evaluation flow
using Innovus 21.1 (see Figure~\ref{fig:evalflow}). 
Table~\ref{tab:cmp} gives postRouteOpt PPA metrics 
for the three macro placement solutions, which are qualitatively
very similar. We observe that CMP 21.1 produces the best rWL result, 
while CMP 19.1 produces the best TNS result.

\begin{table}[]
\caption{Macroplacement solutions generated 
for Ariane-ASAP7 using CMP
in Innovus versions 19.1, 20.1, and 21.1.}
\label{tab:cmp}
\vspace{-0.05in}
\resizebox{\columnwidth}{!}{%
\begin{tabular}{|c|ccccc|cccc|}
\hline
\multirow{2}{*}{\begin{tabular}[c]{@{}c@{}}Macro \\ Placer\end{tabular}} &
  \multicolumn{5}{c|}{PostRouteOpt PPA  (From Innovus)} &
  \multicolumn{4}{c|}{Proxy Cost Details} \\ \cline{2-10} 
 &
  \multicolumn{1}{c|}{\begin{tabular}[c]{@{}c@{}}Area \\ ($\mu m^2$)\end{tabular}} &
  \multicolumn{1}{c|}{\begin{tabular}[c]{@{}c@{}}rWL \\ ($\mu m$)\end{tabular}} &
  \multicolumn{1}{c|}{\begin{tabular}[c]{@{}c@{}}Power \\ ($mW$)\end{tabular}} &
  \multicolumn{1}{c|}{\begin{tabular}[c]{@{}c@{}}WNS \\ ($ps$)\end{tabular}} &
  \begin{tabular}[c]{@{}c@{}}TNS \\ ($ns$)\end{tabular} &
  \multicolumn{1}{c|}{WL} &
  \multicolumn{1}{c|}{Den.} &
  \multicolumn{1}{c|}{Cong.} &
  Proxy \\ \hline \hline
CMP 19.1 &
  \multicolumn{1}{c|}{16407} &
  \multicolumn{1}{c|}{879781} &
  \multicolumn{1}{c|}{504} &
  \multicolumn{1}{c|}{-131} &
  -128.6 &
  \multicolumn{1}{c|}{0.104} &
  \multicolumn{1}{c|}{1.256} &
  \multicolumn{1}{c|}{1.308} &
  1.386 \\ \hline
CMP 20.1 &
  \multicolumn{1}{c|}{16423} &
  \multicolumn{1}{c|}{884527} &
  \multicolumn{1}{c|}{504} &
  \multicolumn{1}{c|}{-155} &
  -196.8 &
  \multicolumn{1}{c|}{0.103} &
  \multicolumn{1}{c|}{1.295} &
  \multicolumn{1}{c|}{1.333} &
  1.417 \\ \hline
CMP 21.1 &
  \multicolumn{1}{c|}{16350} &
  \multicolumn{1}{c|}{843757} &
  \multicolumn{1}{c|}{504} &
  \multicolumn{1}{c|}{-124} &
  -146.1 &
  \multicolumn{1}{c|}{0.102} &
  \multicolumn{1}{c|}{1.122} &
  \multicolumn{1}{c|}{1.141} &
  1.233 \\ \hline
\end{tabular}%
}
\end{table}

\begin{figure*}[!htb]
    \centering
    \includegraphics[ width=1.95\columnwidth]{FIGS/ISPD22_FIGURE8.pdf}
    \vspace{-0.2in}
    \caption{The impact of pre-training versus training from scratch on 
    performance (higher placement return on the y-axis is better) 
    for (a) Ariane-ASAP7, (b) Ariane-GF12 and (c) Ariane-NG45.
    Here, the placement return is defined as the negative of the proxy cost.
    }
    \label{fig:ispd22fig8}
    \vspace{-0.25in}
\end{figure*}

\section{Studies of Pre-training}
\label{sec:pretrain}
In \cite{ispd23}, we do not address pre-training because CT
did not provide pre-training scripts. \cite{YuSJBGMG22} and
\cite{Ariane-pre-train} show that training from scratch
achieves results similar to pre-training. The {\em
Nature} paper does not study the impact of pre-training on PPA
metrics. However, several recent writings \cite{GoldieMYJSWLJPNPTSHTLLHCD24}
\cite{GoldieMD24} \cite{StatementRL} highlight the lack of 
pre-training in~\cite{ispd23}. 
Here, Google's recent open-sourcing of a recipe for 
pre-training~\cite{CT-TCAD}  as well as pre-trained AlphaChip 
model weights ({\em CT-AC})~\cite{CT-TCAD}
enable us to study pre-training in detail.
We are guided by~\cite{YuSJBGMG22} \cite{CT-TCAD}, where
{\em Nature} authors describe two ways to leverage pre-training 
so as to improve fine-tuning results for a given target netlist. 
(i) The first way is to use Google's published AlphaChip checkpoint, 
which is pre-trained on 20 diverse TPU blocks \cite{CT-TCAD}; we refer
to this as {\em CT-AC}.
(ii) The second way is to pre-train a model using similar slices 
of the target block, following guidance from~\cite{YuSJBGMG22} 
\cite{CT-TCAD}; we refer to this as {\em CT-Ours}.

In the following, the first subsection studies 
pre-training on variants of target slices, and fine-tuning on 
the original target slice. This aims to validate ``pretraining on 
different slices of the same block'' from \cite{YuSJBGMG22}. 
The second subsection studies pre-training on variants of slices of 
the target block itself, and fine-tuning on the entire block,
to help with convergence. This aims to use the pre-trained model 
to improve CT scalability on the previously diverged large 
testcase, {\em CT-Ariane-X4}. 
The third subsection studies sensitivity to netlist 
diversity, as well as inherent scalability, of pre-training. 
Here, our studies show limitations of the pre-training 
recipe published in \cite{CT-TCAD}.

\subsection{Pre-training and fine-tuning on slices}
\label{subsec:pre-train-slice}
We study pre-training with diverse variants,
using the Ariane testcase in three technologies: ASAP7, GF12
and NG45. The variants are generated using a perturbation strategy 
adapted from \cite{YuSJBGMG22} (cf. Section 3.2.2 in \cite{YuSJBGMG22}),  
where ``we pre-trained a model on the first 7 slices, and
fine-tune on the 8$^{th}$ slice''.
In our experiment, all slices have the same internal and I/O 
logic, but differ in their I/O port locations.
We apply three operators to produce variant slices
for a given target block:  (i) {\em X-flip} (flip along the x-axis); 
(ii) {\em Y-flip} (flip along the y-axis); and (iii) {\em Shift} (move 
each  I/O clockwise on the canvas boundary, by a distance equal to 
3\% of the length of the canvas side on which the I/O is located).

Given a target netlist with placed I/O ports, we produce 
the 7 slices of the pre-training dataset using 
(i) Shift; (ii) X-flip; (iii) Y-flip;
(iv) XY-flip (flip along both axes); (v) Shift-X-flip (X-Flip
followed by Shift); (vi) Shift-Y-flip (Y-Flip followed by Shift);
and (vii) Shift-XY-flip (XY-Flip followed by Shift).
Then, (viii) the 8$^{th}$ slice is the (unchanged) target itself. 

For each of the Ariane-ASAP7, Ariane-GF12 and 
Ariane-NG45 testcases (each having 133 macros),
we pre-train a model using the first 7 slices.  
As recommended by~\cite{YuSJBGMG22} \cite{CT-TCAD}, 
we use 252 collect jobs (36 collect jobs for 
each slice) to pre-train the model for 200 iterations.
We then run three experiments with the 8$^{th}$ slice (i.e., the 
original target without any change): (i) training {\em CT} from 
scratch; (ii) fine-tuning the {\em CT-Ours} model that we 
pre-trained on the first 7 slices; and (iii) fine-tuning 
Google's public {\em CT-AC} model.

Figure~\ref{fig:ispd22fig8} illustrates the performance 
of these three models -- in terms of placement return versus training time --
for each testcase. 
The model pre-trained from scratch takes similar time to 
converge to a high-quality solution as do the pre-trained 
models. This is qualitatively different from Figure 8 
in \cite{YuSJBGMG22}, where ``starting from scratch takes 5x longer 
to reach a high-quality placement''.
Also, the model fine-tuned using {\em CT-Ours} converges 
to a worse placement compared to the model fine-tuned using 
{\em CT-AC}; this aligns with the observation in \cite{YuSJBGMG22}.

\subsection{Pre-training on slices and fine-tuning on the whole block}
\label{subsec:pre-train-whole}
As shown in Table~\ref{tab:ct_ariane}, {\em CT-Ariane-X4} diverges 
when training from scratch.
We now explore if pre-training 
on {\em CT-Ariane} and {\em CT-Ariane-X2} variants 
helps CT-Ariane-X4 converge during fine-tuning.
Because {\em CT-Ariane-X4} can be constructed 
by replicating {\em CT-Ariane} four times or 
{\em CT-Ariane-X2} twice, we treat {\em CT-Ariane} 
and {\em CT-Ariane-X2} as slices of {\em CT-Ariane-X4},
again following guidance from \cite{YuSJBGMG22}.
Then, we generate six variant slices -- CT-Ariane-X1-\{Shift, X-flip, Y-flip\},
CT-Ariane-X2-\{Shift, X-flip, Y-flip\} -- for pre-training.
With these six slices, we pre-train a model using
252 collect jobs (42 collect jobs per slice) for 200 iterations.

We run three experiments with the original {\em CT-Ariane-X4}:
(i) training {\em CT} from scratch; 
(ii) fine-tuning the {\em CT-Ours} that we pre-trained 
on six {\em CT-Ariane} and {\em CT-Ariane-X2} slices; 
and (iii) fine-tuning Google's public {\em CT-AC} model.
Each fine-tuning experiment is run for 400 iterations. 

\begin{figure}[!htb]
    \centering
    \vspace{-0.1in}
    \includegraphics[ width=0.77\columnwidth]{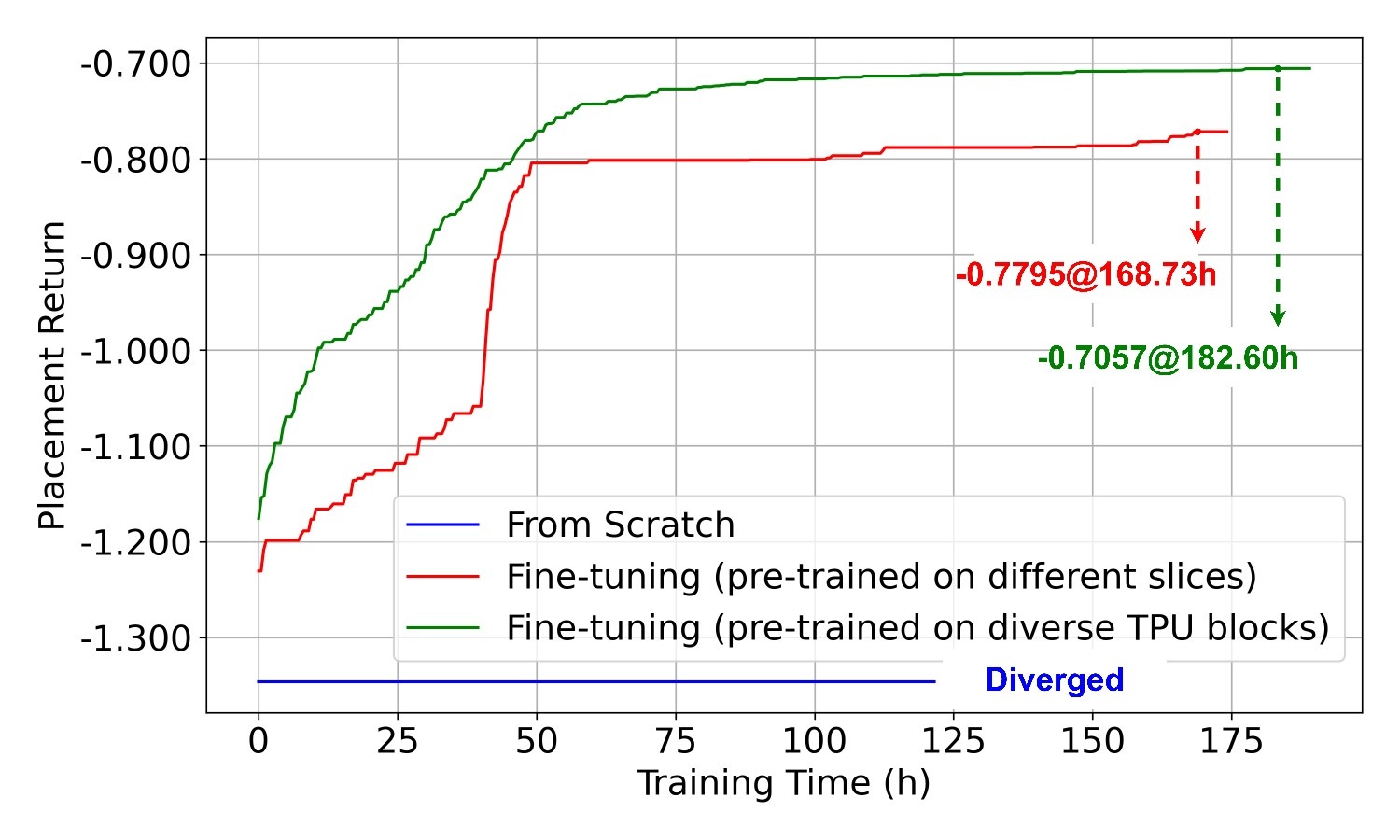}
    \vspace{-0.2in}
    \caption{Placement returns achieved during training 
    from scratch and during fine-tuning from two 
    pre-training strategies: one using data from slices 
    ({\em CT-Ariane} and {\em CT-Ariane-X2}), and the 
    other using a diverse set of TPU blocks. Although the 
    model pre-trained on slices produces a sharp increase 
    in return at 50 hours, it does not outperform the model 
    pre-trained on a diverse set of TPU blocks.
    Here, the placement return is the 
    negative of the proxy cost.}
    \label{fig:x1x2x4}
    \vspace{-0.15in}
\end{figure}

Figure~\ref{fig:x1x2x4} shows placement return versus training
time for our study. We observe the following.
(i) Pre-training enables AlphaChip to converge on a 
larger netlist for which from-scratch training fails.
(ii) The {\em CT-AC} model pre-trained on 20 diverse TPU 
blocks ultimately achieves a higher placement return than 
the {\em CT-Ours} model pre-trained on other 
slices; this is consistent with the observation in 
Subsection \ref{subsec:pre-train-slice}.
(iii) The model pre-trained on other slices sees a 
sharp placement return increase at 50 hours but improves
by only 4\% further over the next 125 hours.
(iv) For {\em CT-Ariane-X4}, the {\em CT-AC} model 
pre-trained on diverse TPU blocks has a lower starting 
point, unlike what we observe in Figure~\ref{fig:ispd22fig8}. 

\subsection{Convergence of pre-training}
\label{subsec:pre-train-diversity}

We study how pre-training convergence relates to the diversity of the 
pre-training dataset and to the size (i.e., \#macros) of the largest 
slice in the pre-training dataset.
Our results show limitations of the current pre-training 
recipe provided in \cite{CT-TCAD}. 

\noindent
\textbf{Diversity.} We use a larger testcase, MemPoolGroup,
to study the impact of pre-training dataset diversity. 
We generate seven slices for MemPoolGroup as described in 
Subsection \ref{subsec:pre-train-slice}. 
From these seven slices, we create five pre-training datasets that are respectively
comprised of 7, 5, 5, 4 and 3 distinct slices. These pre-training datasets
are summarized as Set IDs 1-5 in Table~\ref{tab:PT_diversity}.
For each pre-training dataset, we pre-train a model with at least 252 collect 
jobs (distributed across slices) for 50 iterations.
The rightmost column of Table \ref{tab:PT_diversity} indicates
that higher diversity leads to divergence.\footnote{In the
same pre-training study performed with MemPoolGroup-GF12, all five pre-training
jobs diverge.}

\noindent
\textbf{Scalability.}
To investigate the effect of the number of macros, we use the unperturbed netlists 
of {\em CT-Ariane}, {\em CT-Ariane-X2}, and {\em CT-Ariane-X4} as individual 
pre-training sets, each containing a single netlist. We again use 252 collect 
jobs, and all jobs collect samples on the same unperturbed netlist. 
The last three rows of Table~\ref{tab:PT_diversity} show that pre-training on 
smaller netlists ({\em CT-Ariane} and {\em CT-Ariane-X2}) converges, while 
pre-training on {\em CT-Ariane-X4} diverges.

\noindent
Our diversity and scalability studies of pre-training motivate
several open questions for future investigation, e.g.: 

\begin{itemize}[noitemsep, topsep=0pt, leftmargin=*]
    \item  How much additional computational resource, and/or additional
    iterations, are needed to ensure convergence of pre-training with 
    increased diversity of the pre-training dataset?

    \item When generating slices \cite{YuSJBGMG22}, by how much should 
    each perturbation differ from the original (target) netlist?  (Can 
    too-small or too-large perturbations harm convergence?)
\end{itemize}

\begin{table}[ht]
\vspace{-0.1in}
\caption{Convergence of pre-training according to pre-training set diversity 
and number of macros.}
\vspace{-0.1in}
\centering
\resizebox{0.49\textwidth}{!}{%
\begin{tabular}{|c|c|c|c|c|c|}
\hline
Design  & \begin{tabular}[c]{@{}c@{}}Set \\ ID\end{tabular} 
& Pre-Training Dataset  
& \begin{tabular}[c]{@{}c@{}}Max \\ \#Macro\end{tabular} 
& \begin{tabular}[c]{@{}c@{}}\#Collect\\ Jobs\end{tabular} 
& Converge? \\ \hline
\multirow{5}{*}{\begin{tabular}[c]{@{}c@{}}MemPool\\Group \\ NG45\end{tabular}} 
& 1   
& Shift, \{X,Y,XY\}-flip, Shift-\{X,Y,XY\}-flip
& 324  
& 252   
& No        \\ \cline{2-6} 

& 2 
& Shift, XY-flip, Shift-\{X,Y,XY\}-flip
& 324   
& 255   
& No        \\ \cline{2-6} 

& 3 
& Shift, \{X,Y\}-flip, Shift-\{X,Y\}-flip   
& 324   
& 255                                                      
& No        \\ \cline{2-6} 
                                                                              
& 4                                                 
& Shift, \{X,Y\}-flip, Shift-XY-flip
& 324                                                    
& 256                                                      
& Yes       \\ \cline{2-6} 
                                                                              
& 5                                                 
& Shift, \{X,Y\}-flip                                                            
& 324                                                    
& 252                                                      
& Yes       \\ \hline

\hline

CT-Ariane 
& 6                                                
& Original (no change)                                                                          
& 133                                                    
& 252                                                      
& Yes       \\ \hline

CT-Ariane-X2                                                                  
& 7                                                
& Original (no change)                                                                         
& 266                                                    
& 252                                                      
& Yes       \\ \hline

CT-Ariane-X4                                                               
& 8                                                
& Original (no change)                                                                        
& 532                                                    
& 252                                                      
& No        \\ \hline
\end{tabular}
}
\vspace{-0.15in}
\label{tab:PT_diversity}
\end{table}

\section{Conclusions}
\label{sec:conclusion}

Google's {\em Nature} paper \cite{MirhoseiniGYJSWLJPNPTSHTLLHCD21}, and the
subsequent
GitHub releases of Circuit Training and its ``AlphaChip'' update
\cite{CT-TCAD}, have drawn broad attention throughout the EDA and IC design 
communities.
To the best of our knowledge, no successful reproduction by others of claims 
in \cite{MirhoseiniGYJSWLJPNPTSHTLLHCD21}
has been published in conferences or journals as of \sk{November 2025}.
Meanwhile, {\em Nature} authors have made updates to {\em CT} during the 2.5 years between 
the commits studied in \cite{ispd23} and in this work; notably, these include 
much-welcomed pre-training recipes and the pre-trained AlphaChip model weights. 
This has motivated our continued efforts toward open, transparent 
implementation and a more rigorous assessment of {\em Nature} and {\em CT}.

In this work, we train Google’s AlphaChip from scratch and fine-tune AlphaChip 
(from the pre-trained checkpoint released in August 2024), for all our testcases.
We strengthen the simulated annealing baseline by incorporating multi-threading 
and a 1994 ``go-with-the-winners'' metaheuristic.
Importantly, we add sub-10nm experimental enablement: (i) {\em CT-Ariane} translated
from Google's protobuf, along with scaled versions; and (ii) porting of our testcases 
to the open-source academic ASAP7 PDK. We also perform
pre-training of {\em CT} following instructions 
in the {\em CT} repo \cite{CT-TCAD} and guided by~\cite{YuSJBGMG22}.

Our updated evaluation reconfirms conclusions of \cite{ispd23}.
{\em SA} and human baselines remain superior to
the latest Alpha-Chip, with statistically significant differences in proxy
cost and postRouteOpt PPA metrics, using substantially fewer resources. 
\sk{Moreover}, studies with {\em scaled} sub-10nm Ariane variants
\sk{reveal further weaknesses of \cite{MirhoseiniGYJSWLJPNPTSHTLLHCD21}} -- 
\sk{in} stability, stochasticity, scalability, and compute and runtime demands.
\cite{GoldieMYJSWLJPNPTSHTLLHCD24} notes that ``In any case, AlphaChip 
has been used in production on blocks with over 500 macros''.
At the same time, our experimental results for {\em CT-Ariane-X4} 
(532 macros, TSMC 7nm) indicate that {\em SA} achieves better results than 
AlphaChip in both post-detailed placement HPWL and proxy cost,
using a fraction of runtime and computing resources.

We draw conclusions from data in experiments performed since the {\em Nature}
publication. (i) As baselines should use best available prior
algorithms, using weak or outdated baselines may lead to misleading conclusions.
Careful implementation of strong baselines is crucial for reliable assessments.
(ii) It remains an open research question whether classical metaheuristic methods
will continue to stay ahead of data-hungry AI/ML methods such as RL,
\sk{in large-scale discrete optimizations like macro placement,} where
``ground truth'' is post-P\&R PPA \cite{MirhoseiniGYJSWLJPNPTSHTLLHCD21}.
\sk{Notably, our data show that the proxy cost optimized in {\em CT} correlates
weakly with final post-route PPA metrics (Table~\ref{tab:correlation}),
underscoring a fundamental misalignment between the optimization target used in
{\em Nature} \cite{MirhoseiniGYJSWLJPNPTSHTLLHCD21} and ultimate design objectives.}
(iii) There is no substitute for open access to data and code. Reproducibility 
requires both well-documented methodologies and the exact code implementation. 
(iv) Ideally, a physical design tool should be deterministic, i.e., producing 
the same result given the same machine and parameter settings. If nondeterminism 
exists, variations in results must be clearly acknowledged and analyzed.
And (v) reproducibility is a cornerstone of scientific research. When proposing 
a new methodology, authors should strive to facilitate replication through
comprehensive documentation, results on public benchmarks, and open-source 
code implementation.

The difficulty of reproducing the methods and results 
of \cite{MirhoseiniGYJSWLJPNPTSHTLLHCD21}, and the effort spent on 
{\em MacroPlacement}, highlight the importance of 
``frictionless reproducibility''~\cite{Donoho24}, along 
with open source code and data releases ``upon which others then build'' 
\cite{SpectorN18}, in the academic EDA field and its nexus with AI/ML.
Policy changes of EDA vendors~\cite{Cadence} since late 2022 are a laudable step forward; 
they enable us to include Tcl scripts for commercial synthesis, place and
route flows in the {\em MacroPlacement} GitHub. 
\sk{We are also encouraged by recent research-community interactions
sparked by, e.g., our scaled {\em CT-Ariane} testcases and open-sourced
experimental enablement.} As we wrote in \cite{ispd23}: contributions of benchmarks, 
design enablements, implementation flows and additional studies to the 
{\em MacroPlacement} effort are warmly welcomed.
  
\section*{Acknowledgments}
We thank David Junkin, Patrick Haspel, Angela Hwang and 
their colleagues at Cadence and Synopsys for policy changes that 
permit our methods and results to be reproducible and sharable 
in the open, toward advancement of research in the field.
We thank many Google engineers, in particular
Sergio Guadarrama, Guanhang Wu and Joe Jiang, for 
clarifying many aspects of Circuit Training, and running
their internal {\em CT} flow with our data. We thank the anonymous
reviewers for their diligence, and community members as well as a reviewer
for independently replicating our SA
results. We thank the San Diego Supercomputer Center
for compute resources used in our studies.

\vspace{-0.5in}

\begin{IEEEbiography}[{\raisebox{0.3in}{\includegraphics[height=1.0in,clip,keepaspectratio]
{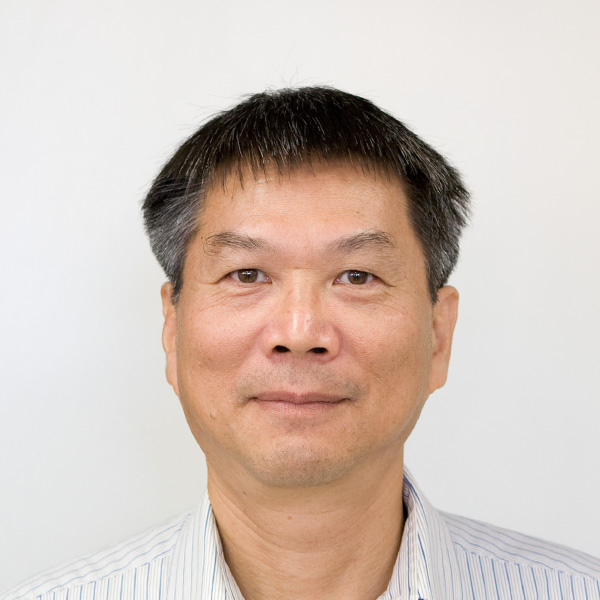}}}]{Chung-Kuan Cheng}
is Distinguished Professor of CSE and Adjunct Professor of
ECE at the University of California, San Diego.
His research interests include machine learning 
and design automation for microelectronic circuits. He received the Ph.D. degree 
in Electrical Engineering and Computer Sciences from the University of
California, Berkeley.
\end{IEEEbiography}

\vspace{-0.7in}

\begin{IEEEbiography}
[{\raisebox{0.4in}{\includegraphics[height=1.0in, clip, keepaspectratio]
{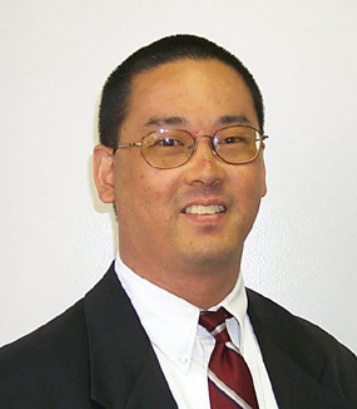}}}]
{Andrew B. Kahng} is Distinguished Professor of CSE and ECE at the 
University of California, San Diego. His interests include IC physical design, 
the design-manufacturing interface, combinatorial optimization, 
and AI/ML for EDA and IC design. He received the Ph.D. degree in Computer Science from the University of California, San Diego.
\end{IEEEbiography}

\vspace{-0.9in}

\begin{IEEEbiography}
[{\raisebox{0.4in}{\includegraphics[height=1.0in, clip, keepaspectratio]{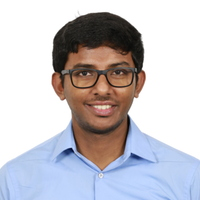}}}]
{Sayak Kundu} is a Ph.D. student in the ECE
department at the University of California,
San Diego. He received his bachelor’s degree in Electronics and Telecommunication Engineering from Jadavpur University, Kolkata, in 2017.
His research interests include optimization and machine learning applications in IC physical design flow and methodology.
\end{IEEEbiography}

\vspace{-0.9in}

\begin{IEEEbiography}
[{\raisebox{0.4in}{\includegraphics[height=1.0in, clip, keepaspectratio]{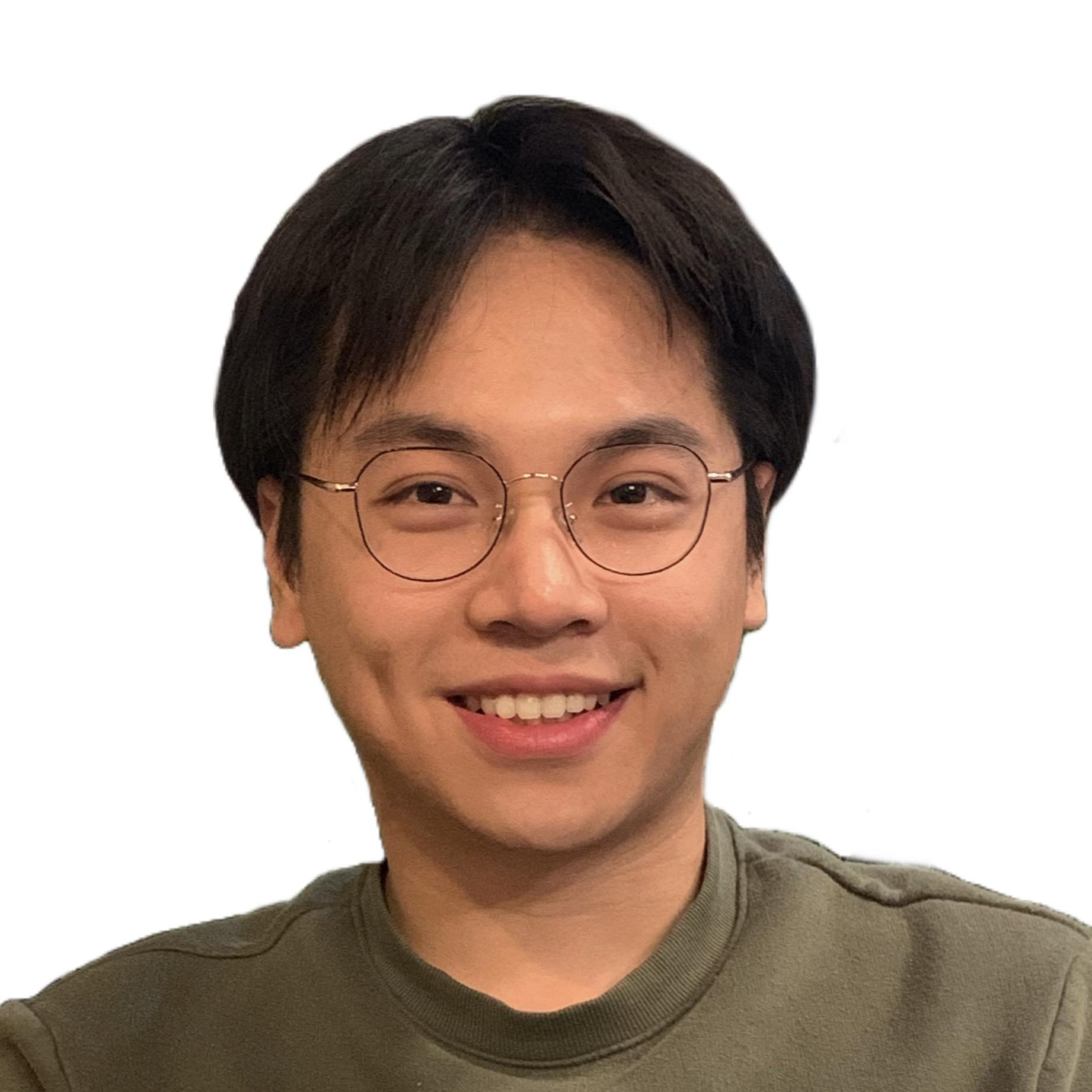}}}]
{Yucheng Wang} is a Ph.D. student in the Computer Science Department of the University of California at San Diego. He received his bachelor's degree in Computer Science from Purdue University, West Lafayette, in 2021. His research interests include standard-cell layout automation, design technology co-optimization strategies, and graph visualization.
\end{IEEEbiography}

\vspace{-0.9in}

\begin{IEEEbiography}
[{\raisebox{0.4in}{\includegraphics[height=1.0in, clip,keepaspectratio]
{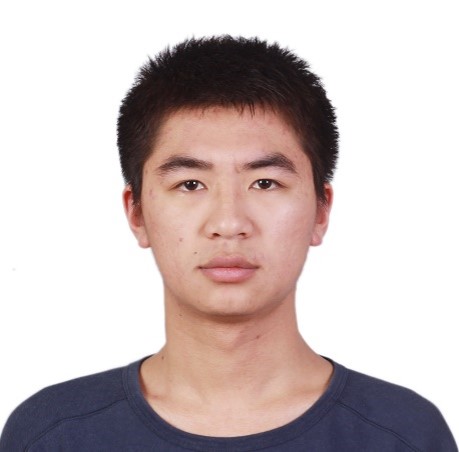}}}]
{Zhiang Wang}
received the Ph.D. degree in electrical and computer
engineering from the University of California, San Diego in 2024. He is currently 
a postdoctoral researcher at the University of California, San Diego. His current 
research interests include physical design, GPU-accelerated EDA and machine learning for EDA.
\end{IEEEbiography}

\appendix
The preceding pages constitute the {\bf early access
version} of  ``An Updated Assessment of Reinforcement 
Learning for Macro Placement'', which is 
our peer-reviewed {\em IEEE Transactions on 
Computer-Aided Design of Integrated Circuits and Systems}
paper.  This appendix provides supplementary information
on journal publication, acknowledgments, additional
experiments, and code/data availability.

\smallskip
\noindent{\bf Journal Publication.}
This v3 revision is the early access version of ``An Updated Assessment 
of Reinforcement Learning for Macro Placement'', published in
{\em IEEE Transactions on Computer-Aided Design of Integrated Circuits 
and Systems}, Digital Object Identifier: DOI 10.1109/TCAD.2025.3644293. 
The early access version, reproduced in the first 14 pages above, is:
\url{https://ieeexplore.ieee.org/stamp/stamp.jsp?arnumber=11300304}.
The final publication version, with IEEE production formatting 
for which we are unable to provide LaTeX sources to meet arXiv
requirements, is available at
\url{https://vlsicad.ucsd.edu/Publications/Journals/j148.pdf}. We refer
to this final publication version as TCAD26.

\smallskip
\noindent{\bf Acknowledgments.}
We thank the anonymous reviewers
for their diligence, and are grateful to community
members as well as one of the reviewers for
independently replicating our results. In particular,
we would like to express our deep appreciation to
Professor Sung-Kyu Lim and his Ph.D. student
MinGyu Park, to Professor Stephan Held, and to
others who have provided independent replication
of SA results that we report in TCAD26.

\smallskip
\noindent{\bf CT-AC-DP Evaluation.}
We use CT-AC-DP to denote the fine-tuning of
AlphaChip starting from the pre-trained checkpoint
released in August 2024, while using DREAMPlace
to place the soft macros.
We conducted the CT-AC-DP experiment at one of the
reviewer's suggestion during the first round of
revision. However, in the second round of
revision, the reviewers ultimately recommended
not including CT-AC-DP results in the final TCAD26 revision.
CT-AC-DP is evaluated on the ASAP7 Ariane
and MemPoolGroup designs, as well as on
the CT-Ariane-X4 design. The details of this
experiment and results are available at:
\url{https://github.com/TILOS-AI-Institute/MacroPlacement/blob/CT-AC-DP/Docs/CT_AC_DP_Results/README.md}.

\smallskip
\noindent{\bf Code and Data Availability.}
All code, scripts and data used in the journal
paper are public in the main branch of
the MacroPlacement repository on GitHub:
\url{https://github.com/TILOS-AI-Institute/MacroPlacement}.
The previous state of the repository is available
in the ``ISPD23'' branch.

\smallskip
\noindent{\bf Addressing Criticisms of ISPD23.}
The TCAD26 paper reports work performed
specifically in response to direct criticisms
leveled at our ISPD23 work~\cite{ispd23},
notably in two sources:
(1)~\cite{GoldieMD24} (``That Chip Has
Sailed''), and
(2)~\cite{StatementRL} (``Statement on
Reinforcement Learning for Chip Design'').
Table~\ref{tab:criticism} summarizes these
criticisms and how TCAD26 addresses each
of them.

\newcolumntype{L}[1]{>{\raggedright\arraybackslash}p{#1}}
\begin{table*}[!t]
\centering
\caption{Summary of criticisms of ISPD23
and how TCAD26 addresses them.
Source~1 = \cite{GoldieMD24};
Source~2 = \cite{StatementRL}.}
\label{tab:criticism}
\scriptsize
\renewcommand{\arraystretch}{1.15}
\begin{tabular}{|L{0.9cm}|L{3.4cm}|L{5.2cm}
|L{5.8cm}|L{1.0cm}|}
\hline
% \rowcolor{gray!15}
\textbf{Source} &
\textbf{Criticism of ISPD23} &
\textbf{TCAD26 Adds/Changes} &
\textbf{Evidence in TCAD26} &
\textbf{Remain-ing Gap} \\
\hline
%% Row A
1, 2 &
A. Did not pre-train the RL method. &
1. Adds explicit pre-training studies
using CT's published instructions.\newline
2. Evaluates fine-tuning from Google's
Aug 2024 pre-trained checkpoint
(``AlphaChip'') in addition to
from-scratch training. &
% Section~\ref{sec:pretrain},
% Table~\ref{tab:PT_diversity}.\newline
Section~\ref{sec:intro} states:
\newline
1. ``Pre-training studies. We perform
pre-training of CT\ldots''\newline
2. ``fine-tune AlphaChip from Google's
August 2024 pre-trained checkpoint.'' &
None \\
\hline
%% Row B
1, 2 &
B. Used an order of magnitude fewer
compute resources. &
1. Used 8 V100 GPUs, batch size 1024.
This is the same GPU resource used in
the ISPD23 paper. (See the quote below
from the Nature authors' ISPD-2022
paper.)
\newline
2. Collect jobs are increased from
26 $\rightarrow$ 256.
\smallskip
\newline
In their ISPD-2022 paper, the Google authors state:
``We think the 8-GPU setup is able to produce better
results primarily because it uses a global batch size
of 1024, which makes learning more stable and reduces
the noise of the policy gradient estimator. Therefore,
we recommend using the full batch size suggested in our
open-source framework [2] in order to achieve optimal
results.”

&
% Section~\ref{sec:replication}.\newline
Section~\ref{subsec:CT_Training} states:
\newline
1. ``Training is conducted on a single
server equipped with eight NVIDIA V100
GPUs, paired with five collect servers,
each featuring a 96-thread CPU''\newline
2. ``Each training job deploys 256
collect jobs, as increasing beyond this
threshold yields diminishing speedup
benefits'' &
None \\
\hline
%% Row C
1, 2 &
C. Did not train to convergence. &
1. Gives CT more opportunity to converge by doubling the number of iterations:
200 $\rightarrow$ 400.\newline
2. Runs multiple trials before declaring
non-convergence. &
% Section~\ref{sec:replication}.\newline
Section~\ref{sec:intro} states:
\newline
1. ``We double iterations from 200 to
400 to provide sufficient opportunity
for CT to converge,''\newline
2. ``and conduct multiple trials before
declaring non-convergence.'' &
None \\
\hline
%% Row D
1, 2 &
D. Evaluated on non-representative,
irreproducible benchmarks. &
1. Releases 7nm Ariane LEF/DEF converted
from protobuf + scaled variants.\newline
2. Evaluates with commercial P\&R to
report post-route PPA on 7nm designs. &
% Tables~\ref{tab:testcase_pretraining}
% and~\ref{tab:ct_ariane}.\newline
Section~\ref{sec:intro} states:
1. ``We convert Google's public TSMC 7nm
Ariane testcase (CT-Ariane) from
protobuf~\cite{CT-TCAD} to LEF/DEF;
we publish this and additional
scaling studies of macro placement
optimizers''\newline
2. ``Studies with these new sub-10nm
enablements reaffirm findings
of~\cite{ispd23}'' &
None \\
\hline
%% Row E
1, 2 &
E. Performed ``massive
reimplementation'' of our
[Nature] method. &
1. Reimplemented {\em plc\_client} is not part
of the main evaluation flow. &
Section~\ref{subsec:comparison} states:
\newline
1. ``Table~\ref{tab:mp_result} also
reports CT proxy cost
for all macro placement solutions, as
evaluated by CT's {\em plc\_client}. To compute
proxy cost for CMP, RePlAce, SA, and
human-expert solutions, we first update
hard macro locations and orientations,
then run FD placement via the {\em plc\_client}
to place all standard-cell clusters
(soft macros).'' &
None \\
\hline
%% Row F
1, 2 &
F. Absence of peer-review [in intro, etc.] &
1. Our TCAD26 paper went through two
rounds of detailed peer review, much
like any other TCAD paper. &
1. The final publication version includes the following dates:
``Received 18 March 2025; revised
13 September 2025 and 11 November 2025;
accepted 6 December 2025.'' &
None \\
\hline
%% Row G
2 &
G. ``RePlAce was state-of-the-art when
we published the Nature paper (and arXiv
preprint). We outperformed it then and
we outperform it now, as shown in the
ISPD paper's Table 1, even ignoring all
of the above issues\ldots'' &
1. TCAD26 points out (Section~II, Page~3,
left column, Line~4) the versions used
in ISPD23, Nature, and TCAD26. Nature
used a version that was deprecated in
January 2021. &
Section~\ref{sec:methods} states:
\newline
1. ``RePlAce~\cite{ChengKKW18}
\cite{RePlAce} models the layout and
netlist as an electrostatic system.
Instances are modeled as electric
charges, and the density penalty as
potential energy. Instances are spread
apart according to the gradient with
respect to the density penalty. Note
that our present work uses RePlAce
from
OpenROAD~\cite{OpenROAD-Ajayi-DAC19}
\cite{RePlAce}, commit hash f02a3d4
from August 2024, which is the
appropriate comparison; our previous
work~\cite{ispd23} used a specific
standalone RePlAce chosen to match the
`Stronger Baselines'
study~\cite{RePlAce2}, and Nature used
a standalone RePlAce from the OpenROAD
project
repository~\cite{PeerReviewNature}
~\cite{CT-RePlAce}, which was
deprecated in January 2021.'' &
None \\
\hline
%% Row H
2 &
H. ``Although the study is entitled
`Assessment of Reinforcement Learning
for Macro Placement', it does not
compare against or even acknowledge any
of the recent RL methods building on our
work (link).'' &
1. In TCAD26 (with concurrence of editor
and reviewers), our focus remained on
assessment of the Nature paper. &
1. The first sentence of the TCAD26
abstract states:
``We provide an improved assessment of
Google Brain's deep reinforcement
learning approach to macro placement
(Nature, 2021) and its updated Circuit
Training (CT) implementation in
GitHub.'' &
None \\
\hline
%% Row I
2 &
I. ``\ldots the ISPD paper compares CT
against AutoDMP (ISPD 2023) and
(presumably) the latest version of CMP,
a black-box, closed-source commercial
autoplacer. Neither of these methods
were available when we released our
paper in 2020.'' &
1. CMP was present in Innovus 19.1, 20.1
which predate the publication of the
Nature paper. The TCAD26 paper shows
CMP results from Innovus 19.1,
20.1 and 21.1 for Ariane on ASAP7.
\newline
2. The TCAD26 paper drops AutoDMP
results from the
main comparison. &
1. Table~\ref{tab:cmp} of the TCAD26
paper and
corresponding
text.\newline
2. Footnote~\ref{fn:two} of the TCAD26
paper specifically notes: ``In this
work, we do not
compare with AutoDMP and Hier-RTLMP, as
these were released after the Nature
work.'' &
None \\
\hline
%% Row J
2 &
J. ``focuses on the use of the initial
placement from physical synthesis to
cluster standard cells, but this is of
no practical concern\ldots'' &
1. For completeness, the TCAD26 paper
performed a similar ablation study
as in the ISPD23 paper,
using the August 2024 CT-AC version. &
% Table~\ref{tab:init_placement}.\newline
Section~\ref{subsec:ablation} states:
\newline
1. ``This observation differs from that
of~\cite{ispd23}, which found up to 10\%
improvement in rWL when a Genus iSpatial
placement solution was used.
(\cite{ispd23} used the older version of
CT, trained for 200 iterations, and
tested on the Ariane-NG45 design.)'' &
None \\
\hline
\end{tabular}
\end{table*}

\end{document}